\documentclass[10pt,twocolumn,letterpaper]{article}

\usepackage[pagenumbers]{cvpr}

\usepackage{booktabs}
\usepackage{graphicx}
\usepackage{xspace}
\usepackage{amsmath}
\usepackage{amssymb}
\usepackage{multirow}

\DeclareMathOperator*{\argmin}{arg\,min}

\newcommand{\paper}{SNAP\xspace}
\newcommand{\snaphead}{SnapHead\xspace}

\definecolor{cvprblue}{rgb}{0.21,0.49,0.74}
\usepackage[pagebackref,breaklinks,colorlinks,allcolors=cvprblue]{hyperref}

\def\paperID{*****}
\def\confName{CVPR}
\def\confYear{2028}

\title{Adapting Frozen Promptable Segmenters to User-Defined Taxonomies via Few-Shot Proposal Learning}

\author{Khayrul Islam, Christopher Frederic Miller\\[2pt]
\normalsize Computational Engineering Division, Lawrence Livermore National Laboratory,\\
\normalsize 7000 East Avenue, Livermore, CA 94550, USA}

\newcommand{\llnlrelease}{LLNL-CONF-2026172}

\begin{document}
\maketitle
{\let\thefootnote\relax\footnotetext{This work was performed under the auspices of the U.S.\ Department of Energy by Lawrence Livermore National Laboratory under Contract \mbox{DE-AC52-07NA27344}. Release number: \mbox{\llnlrelease}.}}
\begin{abstract}
Promptable segmenters of the SAM family turn a click or a phrase into an instance mask, so the costly part of a label is now its class.
However, each annotation project draws its own class boundary, often through objects these models already know: people from cars, sedans from trucks, one sedan model from another.
A generic model's boundary is fixed by its training vocabulary.
We present \paper, an annotation pipeline that learns the project's boundary from the labels the annotator draws anyway.
On one frozen SAM2.1 pass per image, we fit a per-project head, \snaphead, that classifies every proposal and refits in seconds after each labeled image.
Where SAM~3 is available, the same labels decide class by class whether to keep its boundary or ours.
Given the true masks of eight look-alike cell lines, SAM~3 assigns classes at Cohen's $\kappa = 0.038$ and \snaphead at $0.754$ ($60$ labeled images), and when only the class names change, SAM~3 falls to chance and \snaphead's score is unchanged.
End to end, with $20$ labeled images, \paper beats a SAM~3 oracle tuned on the evaluation labels on five microscopy projects ($0.484$ against $0.441$ F1), and run unchanged on five held-out datasets, it beats SAM~3 as shipped on four.
Our results suggest that frozen foundation features already encode the class boundaries a generic vocabulary misses, and that a project's own labels are enough to read them out.
\end{abstract}
\begin{figure}[t]
\centering
\includegraphics[width=\columnwidth]{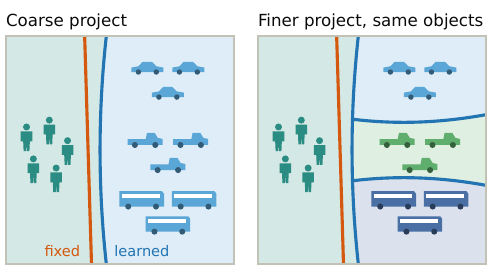}
\caption{A generic segmenter has a fixed class boundary, but every project needs its own. Two annotation projects over the same scene, colored by class. SAM~3's boundary (orange) is fixed by its training vocabulary and stays at the coarse person/car cut. A boundary learned from a project's own labels (blue) moves to the finer sedan/truck/bus cut that project needs.}
\label{fig:teaser}
\end{figure}

\section{Introduction}
\label{sec:intro}

Instance masks with class labels are the slowest part of building a vision dataset.
Promptable segmenters such as SAM, SAM2 and SAM~3~\cite{kirillov2023sam,ravi2025sam2,carion2026sam3} made the mask itself cheap: a click, a box or a short phrase returns a mask with no training on the project.
A label, however, is a mask plus a class, and the class is now the costly half.
Annotation tools put one public checkpoint behind every prompt~\cite{cvat,tkachenko2020labelstudio,roboflow}, so the class comes either from the annotator, one object at a time, or from the model's own training vocabulary.
In the first case, cost still grows with the number of objects; in the second, the classes are limited to the model's training vocabulary.

Where a project draws its class line depends on the project.
Separating people from cars is a cut SAM~3 already makes.
Sedans versus trucks and buses is finer, one sedan model against another is finer still, and one brand against another is finer again.
The annotator knows exactly which cut they want; a generic model was never trained to make most of them.
Our key observation is that an annotation project rarely brings new objects; it brings a new boundary between objects a model has already seen (\cref{fig:teaser}).
The gap persists when we supply the correct masks and ask for the class alone: SAM~3 tells common animal species apart almost perfectly, yet on eight look-alike cell lines it does no better than always naming the most common line.
Better words do not close the gap: SAM~3 finds nuclei from ``bright spots'' but almost none from ``nucleus'', and without labels only the answer key tells which phrase works (\cref{sec:master}).
A failed prompt gives the model nothing to learn from, while every label the annotator draws is a training example.

A project can already get its own boundary, at a price.
Fine-tuning a detector on the project's labels~\cite{he2017maskrcnn,madan2024revisiting}, or retraining a network each round as labels arrive~\cite{diazpinto2024monai}, fits any cut, but every project, and often every round, pays for a full training run.
Pixel classifiers on frozen features~\cite{sommer2011ilastik,berg2019ilastik,seifi2025featureforest} train in seconds, but they label pixels, so the annotator is still left to split and check each instance.
What is missing is a way to learn an instance-level boundary from the few images the annotator has already labeled, fast enough to refit after every image and with nothing set by hand for each project.

We therefore keep the segmenter frozen and learn only the part that changes between projects, the class boundary.
\paper (\cref{sec:method}) runs one procedure, unchanged, on every project.
Class-agnostic proposers find candidate masks on the features of one frozen SAM2.1 pass~\cite{ravi2025sam2}, and each candidate is described by a vector read from the same encoder pass that produced its mask.
A small per-project head, \snaphead, fit to the annotator's labels in seconds, gives each candidate a class or rejects it as background.
SAM2's automatic mask generator supplies proposals from the first image; once labels exist, a proposer trained on the cached features joins it where the generator's point grid misses densely packed objects, and the labels decide when.
Proposals from other sources enter the same path, and when SAM~3 is one of them, the labels also decide, class by class, whether its generic boundary or the learned one is kept, so a stronger generic model adds to the learned boundary.
Each component builds on established tools; together they give a class boundary set by each project's own labels, refit in seconds after every labeled image, unchanged when only the class names change (last two rows of \cref{tab:naming}), and weighed class by class against SAM~3's own boundary on the same labels.

We judge the tool by the work it leaves the annotator.
Our main measure is the drawn fraction, which needs no cost model; for total effort, priced from published timing studies, we report it over the whole cited range.
On every development project, the annotator draws fewer objects by hand with \paper than without it.
Run once, frozen, on held-out datasets that played no part in its design, it beats SAM~3 as shipped on most of them.
On cuts outside SAM~3's vocabulary, such as look-alike cell lines, it beats SAM~3 even with prompt and threshold chosen on the evaluation labels (\cref{sec:experiments}).
Where the cut already sits in SAM~3's vocabulary, the labels hand most classes to SAM~3, and once the cut moves past it they hand them to \snaphead; on microscopy this combination beats SAM~3's oracle.

In summary, we make three contributions:
\begin{enumerate}\itemsep1pt
\item \textbf{\texttt{Learned Boundary}}: \snaphead, fit in seconds to the annotator's labels, gives every proposal a class or background from a descriptor read from the same SAM2.1 pass that produces its mask (\cref{sec:head}).
\item \textbf{\texttt{Plug-in Proposals}}: Any class-agnostic source, from SAM2's mask generator to a few-shot proposer on cached features or Cellpose, feeds the same \snaphead; the project's labels decide which sources run and, when SAM~3 is available, whose class boundary each class uses (\cref{sec:proposals,sec:absorption,sec:router}).
\item \textbf{\texttt{Effort Accounting}}: An optional per-class threshold never costs more than drawing by hand on its tuning evidence, and the bound holds on $221/231$ held-out splits; it charges each prediction the annotator action it triggers, priced from published timing studies (\cref{sec:effort}).
\end{enumerate}
\section{Related work}
\label{sec:related}

\paragraph{Promptable and open-vocabulary segmentation.}
SAM~\cite{kirillov2023sam} and SAM2~\cite{ravi2025sam2} turn a click or a box into a mask; their automatic mask generator sweeps a point grid over the image, but neither assigns a class.
SAM~3~\cite{carion2026sam3} adds a concept path (a noun phrase or an image exemplar) that names what to segment, and SAM~3.1 is a checkpoint release without a paper~\cite{meta2026sam31}.
Grounding DINO~\cite{liu2024groundingdino}, YOLOE~\cite{wang2025yoloe} and other text- and exemplar-prompted detectors~\cite{jiang2024trex2,ren2024dinox,cheng2024yoloworld} reach the same interface from detection.
Fine-grained benchmarks such as Stanford Cars~\cite{krause2013cars} and FGVC-Aircraft~\cite{maji2013aircraft} separate objects that differ only in make, model or variant, and open-vocabulary detectors drift as the label moves down such a hierarchy~\cite{liu2024shine}; SAM~3's authors note that fine-grained concepts remain hard for their model~\cite{carion2026sam3}.
In these models the class boundary is set by the training vocabulary; we learn it from each project's own labels.

\paragraph{Few-shot and per-project adaptation.}
A project boundary can also come from fine-tuning.
Few-shot object detectors fine-tune a detector's last layer on a handful of boxes per class~\cite{wang2020frustratingly} or fine-tune a vision-language detector on a few multimodal examples per target class~\cite{madan2024revisiting}, and each new class needs a retrain.
FeatureForest~\cite{seifi2025featureforest} fits a random forest on frozen SAM2 and DINOv2 features from a few user strokes, in the line of pixel classifiers inside bioimage tools~\cite{sommer2011ilastik,berg2019ilastik,argandacarreras2017trainable,arzt2022labkit,hinderling2024convpaint}; we run this recipe as a baseline.
Teuber \etal~\cite{teuber2026evaluating} probe SAM-family embeddings for classification with instance masks given as input, and linear probes on SAM tokens appear in analyses on episodic splits~\cite{kim2024sola,cuttano2025sansa}.
M2C~\cite{zhou2026mask} fits a small concept embedding to a frozen SAM~3 from a few labeled masks for medical semantic segmentation.
Episodic few-shot methods~\cite{zhang2024personalize,liu2024matcher,sun2024vrp} and a spatial-canvas recipe for SAM~3~\cite{tsai2026fewshot} transfer one concept from a support image, and MultiverSeg~\cite{wong2025multiverseg} grows an in-context set~\cite{butoi2023universeg} to segment one structure at a time.
In microscopy, CellSAM~\cite{marks2025cellsam}, micro-SAM~\cite{archit2025microsam} and specialists such as StarDist and Cellpose~\cite{schmidt2018stardist,stringer2021cellpose,pachitariu2025cellposesam} train backbones or decoders on large curated corpora of annotated cells.
SAMCell~\cite{vandeloo2025samcell} fine-tunes SAM to predict a distance map that watershed~\cite{bai2017deepwatershed} decodes into instances.
These methods adapt a model to one concept, label pixels, fine-tune on many boxes per class, or retrain on thousands of masks; we fit \snaphead, a multi-class classifier, together with a proposal stage to the few images the annotator has labeled.

\paragraph{Annotation tools with a model in the loop.}
CVAT~\cite{cvat}, Label Studio~\cite{tkachenko2020labelstudio} and Roboflow~\cite{roboflow} embed SAM-family interactors, as do desktop and bioimage tools~\cite{wang2023xanylabeling,garcialopezdeharo2026samj,dinya2026samannot,oehme2026aramsam,napari}.
In these tools one public checkpoint answers every prompt in every project: the annotator still prompts each object, and cost grows with the object count.
Pre-labeling followed by human correction is equally established~\cite{russell2008labelme,andriluka2018fluid,uijlings2020panoptic}, and Fluid Annotation~\cite{andriluka2018fluid} edits a full-image proposal set with a few actions per object.
Where a project model exists, as in MONAI Label~\cite{diazpinto2024monai} and related tools~\cite{tenckhoff2024feedback,marchesoniacland2023iadet,zhang2022onelabeler}, it is a full network retrained each round, with a confidence slider as the main cost control.
We fit \snaphead in seconds on a frozen encoder, with an explicit background class that lets a proposal be rejected and an operating point that minimizes the annotator's expected effort (\cref{sec:head,sec:effort}).

\paragraph{Proposals and effort evaluation.}
R-CNN and its successors~\cite{girshick2014rich,ren2015faster,he2017maskrcnn} established propose-then-classify.
Annotation cost is commonly counted in annotator actions: vertex corrections~\cite{castrejon2017polygonrnn,acuna2018polygonrnnpp}, clicks to a quality target in interactive segmentation~\cite{xu2016deep,sofiiuk2022ritm}, and actions per object in large annotation campaigns~\cite{benenson2019largescale,kuznetsova2020open}.
Timed studies supply the relative cost of each action, from drawing and refining a mask to verifying a proposal~\cite{tyagi2023boxcell,oehme2026aramsam,andriluka2018fluid,papadopoulos2016verify}.
RClicks~\cite{antonov2024rclicks} shows that click-position oracles overstate interactive performance, a bias our simulated annotator avoids, since it places no clicks.
For choosing which image to label next, standard active-learning strategies~\cite{sener2018coreset,gal2017baldicml} rarely beat random selection at low budgets~\cite{hacohen2022budget,yu2023twostep}.
We build on both lines.
Our learned proposer follows CenterNet's formulation~\cite{zhou2019centernet,law2018cornernet} on cached encoder features, and our effort model counts every action of a simulated annotator (accept, refine, relabel, delete, draw) and prices each one with weights drawn from these timed studies.
\section{Method}
\label{sec:method}

\begin{figure*}[t]
\centering
\includegraphics[width=\textwidth]{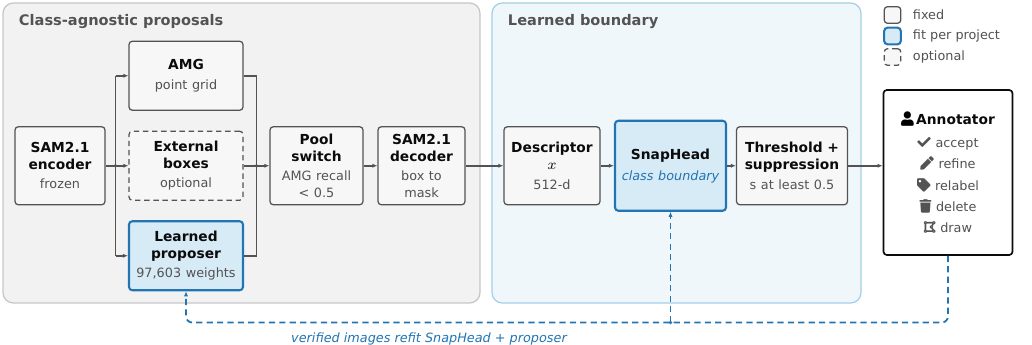}
\caption{Overview of \paper. Gray blocks are frozen, blue blocks are fit to the labeled images $\mathcal{L}$, and the dashed block is an optional proposal source. Every proposal passes through one frozen SAM2.1 decoder, and each newly labeled image refits \snaphead and the learned proposer.}
\label{fig:pipeline}
\end{figure*}

Given a frozen promptable segmenter and the few images an annotator has labeled, our goal is to segment and classify the objects in the rest of the project.
We therefore keep the segmenter frozen and learn two small parts from the labels: a proposer for the objects its point grid misses (\cref{sec:proposals}) and \snaphead, which draws the project's class boundary (\cref{sec:head}).
When SAM~3 is available, the same labels also decide, per class, whether its boundary or \snaphead's is kept (\cref{sec:router}).

\subsection{Problem setup}
\label{sec:formulation}

A project defines a set of classes $\mathcal{C}$ over a pool of images.
Once the annotator has finished $N$ images, the labeled set is $\mathcal{L} = \{(I_j, \{(g_k, y_k)\}_k)\}_{j=1}^{N}$, where $g_k$ is an object mask and $y_k \in \mathcal{C}$ its class.
For every other image $I$ we predict masks with classes, $\{(m_i, \hat{y}_i)\}$ with $\hat{y}_i \in \mathcal{C}$.
We build on SAM2.1~\cite{ravi2025sam2}: its encoder $\Phi$ runs once per image, and its decoder $\Psi$ turns a point or box prompt into a mask $m$, a mask-quality estimate $q$ and an object pointer $o$; both stay frozen.
We fit three parts, all from $\mathcal{L}$: the proposer weights $\theta$, the classifier weights $W$ and the per-class choice of boundary, and refit all three as $\mathcal{L}$ grows.

\subsection{Plug-in proposals}
\label{sec:proposals}

\snaphead can only classify an object that some proposer has found.
SAM2's automatic mask generator (AMG) prompts $\Psi$ with a regular point grid and keeps the masks that pass a quality and overlap filter, which gives a proposal set $P_{\mathrm{AMG}}(I)$.
The grid needs no labels, so it works from the first image, but its fixed spacing misses objects that are small and densely packed.

\paragraph{Learning where the grid fails.}
\label{sec:proposer}
To recover these objects, we fit a second proposer $g_\theta$ on the annotator's own masks.
It follows CenterNet~\cite{zhou2019centernet}: from the stride-4, -8 and -16 features that $\Phi$ already computes, normalized per image, it predicts a centre heatmap $\hat{H}$ and a log-size map $\hat{S}$, trained with
\begin{equation}
\mathcal{L}_{\theta} = \mathcal{L}_{\mathrm{focal}}(\hat{H}, H) + \frac{1}{|\Omega|} \sum_{u \in \Omega} \big\| \hat{S}_u - \log(w_u, h_u) \big\|_1,
\label{eq:proposer}
\end{equation}
where $H$ places a Gaussian at every object centre, $\Omega$ is the set of centre cells, $(w_u, h_u)$ is the box size of the object centred at $u$, and $\mathcal{L}_{\mathrm{focal}}$ is the focal loss of CornerNet~\cite{law2018cornernet}.
The top-$K$ heatmap peaks become boxes, and each box prompts $\Psi$ as an annotator's box would, which gives $P_\theta(I)$.
We set $K$ to $\kappa$ times the median object count of a labeled image, clipped to $[K_{\min}, K_{\max}]$, so a crowded project gets a larger budget than a sparse one.
Since $g_\theta$ reads cached features, it adds no encoder pass.

\paragraph{Choosing the pool from the labels.}
The learned proposer pays off only where the grid fails, and the labels already show where that is.
Let $G$ pool the object masks of all images in $\mathcal{L}$.
We measure the grid's recall on them and switch the pool on it:
\begin{equation}
\begin{aligned}
r &= \frac{1}{|G|} \sum_{g \in G} \mathbf{1}\Big[ \max_{m \in P_{\mathrm{AMG}}} \mathrm{IoU}(m, g) \ge 0.5 \Big], \\
P &= \begin{cases} P_{\mathrm{AMG}} \cup P_\theta & \text{if } r < \tau_r, \\ P_{\mathrm{AMG}} & \text{otherwise.} \end{cases}
\end{aligned}
\label{eq:gate}
\end{equation}
The switch reads only labels the annotator has already drawn, and the grid's proposals stay in the pool either way.

\paragraph{Any proposal source plugs in.}
\label{sec:absorption}
Every proposal reaches \snaphead the same way: as a box prompt to the frozen decoder $\Psi$, followed by its descriptor $x$ (\cref{eq:descriptor}).
The pool is therefore open to any class-agnostic source.
Boxes from a zero-shot detector such as SAM~3~\cite{carion2026sam3}, Grounding DINO~\cite{liu2024groundingdino} or YOLOE~\cite{wang2025yoloe} join $P$ alongside $P_{\mathrm{AMG}}$ and $P_\theta$, with no change to any other part of the pipeline.
Unless stated otherwise, results use $P_{\mathrm{AMG}}$ and $P_\theta$ only; \cref{tab:master} also adds Cellpose~\cite{stringer2021cellpose} boxes to $P_{\mathrm{AMG}}$.

\subsection{Learning the project boundary}
\label{sec:head}

The class boundary is the part that changes between projects, so we read it out of the frozen model instead of retraining it.
For every proposal, $\Psi$ already computes an embedding of the region and an object pointer, so we describe a proposal with mask $m$ by their concatenation:
\begin{equation}
x = \big[\, \bar{\phi}(m) \,;\, o(b(m)) \,\big] \in \mathbb{R}^{512},
\label{eq:descriptor}
\end{equation}
where $\bar{\phi}(m)$ averages the stride-16 embedding of $\Phi$ over $m$, $o(b(m))$ is the object pointer returned for the tight box $b(m)$, and each half is $\ell_2$-normalized.
The first half describes how the region looks; the second is the decoder's own summary of the object it segmented, and from two labeled images on, the pair beats either half alone (\cref{sec:sup_halves}).
Both come from the pass that produced the mask, so the descriptor needs no second model.

\paragraph{Fitting \snaphead.}
The labels supply examples of every class and of background at once.
On every labeled image, a proposal that overlaps an object of class $c$ at IoU $0.5$ or more becomes an example of $c$, and one that overlaps every object below $0.1$ becomes an example of background, $\varnothing$; the ambiguous rest are left out.
On standardized descriptors $\tilde{x}$ we fit a multinomial logistic regression,
\begin{equation}
\min_W \; \sum_i \omega_{y_i}\, \ell_{\mathrm{CE}}\big(\mathrm{softmax}(W \tilde{x}_i),\, y_i\big) + \lambda \|W\|_2^2,
\label{eq:head}
\end{equation}
where $y_i \in \mathcal{C} \cup \{\varnothing\}$ and $\omega_y$ is inversely proportional to the number of examples of $y$, so a rare class weighs as much as a common one.
Linear on a fixed 512-d input, \snaphead fits in seconds, which is what lets it refit after every labeled image.

\paragraph{From probabilities to masks.}
A proposal is kept only if its most likely label is a class rather than $\varnothing$.
It then takes its best class $\hat{c}$ and the score
\begin{equation}
s = \frac{p(\hat{c} \mid x)}{\sum_{c \in \mathcal{C}} p(c \mid x)}\; q,
\label{eq:score}
\end{equation}
the confidence among the named classes times the decoder's own mask quality, so a well-classified but poorly segmented proposal still ranks low.
We keep proposals with $s \ge \tau$, one threshold for every class, and remove duplicates by greedy suppression in score order at mask IoU $\tau_{\mathrm{nms}}$; \cref{sec:sup_ablations} studies a containment-aware variant of this step.

\paragraph{Choosing whose boundary each class uses.}
\label{sec:router}
Where a cut already lies in SAM~3's vocabulary, SAM~3's own class is hard to beat (\cref{sec:naming}).
When SAM~3 is available, we therefore let the labels choose, one class at a time.
On $\mathcal{L}$ we pick SAM~3's phrasing and confidence threshold by macro-F1 from the same five phrasings and seven thresholds its oracle searches (\cref{sec:setup}), and we score \snaphead on $\mathcal{L}$ by five-fold cross-validation grouped by image.
Each class $c$ keeps the source with the higher F1 on its labeled instances, and a class with no labeled instance keeps SAM~3.
On a new image, each class takes its detections from its source, and greedy suppression in score order at mask IoU $\tau_{\mathrm{nms}}$ removes duplicates across sources; no choice reads an evaluation label.

\subsection{Pricing predictions by annotator action}
\label{sec:annotation}

F1 scores a set of predictions but not the work it leaves: a wrong class costs a click to fix, while a missed object costs a full drawing.
We therefore charge each prediction the action it triggers in a simulated annotator, in the spirit of Fluid Annotation~\cite{andriluka2018fluid}.

\paragraph{Effort accounting.}
\label{sec:effort}
Predictions are matched to ground truth greedily in score order at mask IoU $0.5$.
A matched prediction of the right class is an \texttt{accept} if its IoU reaches $\tau_{\mathrm{acc}}$ and a \texttt{refine} otherwise; one of the wrong class is a \texttt{relabel}; an unmatched prediction is a \texttt{delete}; and every object left unmatched is a \texttt{draw}.
Each action $a$ costs $c_a$ in units of one draw: $c_{\mathrm{draw}} = 1$, $c_{\mathrm{refine}} = w_{\mathrm{refine}}$, and the three \emph{check} actions, \texttt{accept}, \texttt{relabel} and \texttt{delete}, each a look and at most one click, cost $w_{\mathrm{check}}$, with $0 < w_{\mathrm{check}} \le w_{\mathrm{refine}} < 1$.
Effort per object is then
\begin{equation}
E(w_{\mathrm{check}}, w_{\mathrm{refine}}) = D + w_{\mathrm{refine}}\,R + w_{\mathrm{check}}\,C,
\label{eq:cost}
\end{equation}
where $D$, $R$ and $C$ are the draw, refine and check shares of the object count.
The drawn fraction $D$ needs no weight, so we lead with it; the two weights come from published timing studies, and we report $E$ over their whole cited range (\cref{sec:setup}).

\paragraph{Effort-optimal thresholds.}
A single $\tau$ treats an easy and a hard class alike, and the same accounting can give each class its own operating point.
We decompose the total cost of all actions over classes as
\begin{equation}
A(t) = n_{\mathrm{gt}}\,c_{\mathrm{draw}} + \sum_c \delta_c(t_c),
\label{eq:effort}
\end{equation}
where $\delta_c(t)$ sums, over the detections of class $c$ kept at threshold $t$, the cost $c_a$ of each triggered action minus one $c_{\mathrm{draw}}$ per ground-truth object claimed.
The optional per-class dial sets $t_c = \argmin_t \delta_c(t)$ on cross-validated curves over $\mathcal{L}$, grouped by image.
Each curve includes the choice of dropping class $c$ entirely, so $\delta_c \le 0$ at the optimum and $A(t) \le n_{\mathrm{gt}}\,c_{\mathrm{draw}}$ for any choice of costs: on its tuning evidence, the dial never costs more than drawing every object by hand.
The flat $\tau$ is one point on the same curves, and \cref{sec:sup_guarantee} measures how far the bound carries to unseen images.
\section{Experiments}
\label{sec:experiments}

\subsection{Setup}
\label{sec:setup}

\paragraph{Projects and splits.}
Thirteen public projects form our development set: six natural, two aerial and five microscopy.
The natural set is COCO~\cite{lin2014coco} vehicles, kitchen and animals, LVIS-rare20~\cite{gupta2019lvis}, VOC-20~\cite{everingham2010pascal,everingham2015pascal} and ADE20K-indoor~\cite{zhou2019ade20k}; the aerial set is iSAID~\cite{zamir2019isaid} vehicles and buildings on DOTA~\cite{xia2018dota} tiles; the microscopy set is LIVECell~\cite{edlund2021livecell} A172, SH-SY5Y, the eight-cell-line mix, the LIVECell split defined by micro-SAM~\cite{archit2025microsam} and BBBC038~\cite{caicedo2019dsb,ljosa2012bbbc} nuclei.
Each project has 180 images: a pool of 60 from which the annotator labels $N \in \{1,2,5,10,20,50\}$ images, and 120 held out for evaluation.
We report mean $\pm$ sd over three seeds.
Five further datasets are held out for transfer (\cref{sec:transfer}).
After fixing the configuration of \cref{tab:config}, we drew them from RF100-VL~\cite{robicheaux2025rf100vl}, a benchmark new to this work: we sorted the $74$ datasets with at least two classes and $180$ images by name and used the first five of a random permutation as drawn.
\paragraph{Baselines.}
All methods in \cref{tab:master} use the same split; our variants also share the encoder and \snaphead and differ only in the proposal source or, for the router, in whether a class keeps SAM~3's own label.
Our strongest baseline is SAM~3~\cite{carion2026sam3} with an oracle choice of phrasing and threshold.
We prompt it with five distinct phrasings of every class: the class name, a synonym, a hypernym, the name with a true modifier (``cell'' $\rightarrow$ ``phase contrast cell''), and a short description of what the object looks like (``nucleus'' $\rightarrow$ ``bright spots'').
Each phrasing is scored at the best of seven confidence thresholds, chosen on the evaluation labels.
The best of the five is therefore an oracle unavailable in practice, and the strongest prompt-only baseline.
We also report the expected F1 of a blind pick among the five.
Grounding DINO~\cite{liu2024groundingdino} and YOLOE~\cite{wang2025yoloe}, given the same five phrasings and each its own oracle, trail SAM~3 on 12 of the 13 projects; on the eight cell lines all three score near zero (\cref{tab:detectors}).
The PerSAM baseline uses no text: it matches a prototype from one support example against the same frozen proposals, which scores above PerSAM, PerSAM-F and Matcher at their published settings (\cref{sec:sup_exemplar}).
The pixel classifier is a random forest on the same frozen SAM2 features followed by connected components, the ilastik and FeatureForest recipe~\cite{berg2019ilastik,seifi2025featureforest}, trained on the same labeled images.
Few-shot Mask~R-CNN~\cite{he2017maskrcnn} is trained on the same labels at all six budgets, including $N{=}1$.
The manual baseline draws every object from scratch.

\paragraph{Metrics.}
Our main metric is instance macro-F1 at mask IoU $0.5$, with greedy matching in score order; we also report AP$_{50}$.
For the proposer we report class-agnostic recall at IoU $0.5$ and $0.75$, averaged over images.
For annotation cost we report the drawn fraction $D$, the share of objects the annotator still draws from scratch, together with corrections per object and the effort $E(w_{\mathrm{check}}, w_{\mathrm{refine}})$ of \cref{sec:effort} (Eq.~\ref{eq:cost}).
Timed studies set the range of both weights: verifying a box takes $1.6$\,s against $26$--$42$\,s to draw it~\cite{papadopoulos2016verify} and $3.5$\,s against $7.4$\,s~\cite{kuznetsova2020open}, and assisted refinement takes $2.6$\,s against $9.7$\,s~\cite{oehme2026aramsam} and $4.20$\,s against $9.92$\,s~\cite{tyagi2023boxcell}.
We mark defaults $w_{\mathrm{check}} = 0.1$ and $w_{\mathrm{refine}} = 0.5$, the latter above both refinement ratios, a setting that favours the baselines, and report $E$ over the whole cited range.
One configuration (\cref{tab:config}) serves every project, and code and configurations will be released on GitHub.
\subsection{Fine-grained class assignment}
\label{sec:naming}

To separate class assignment from localization, we give every method the same ground-truth instance masks and score only the assigned class.
SAM~3 uses its prompt oracle: the best of five phrasings and four decision rules, chosen on the evaluation labels, at confidence threshold $0$.
\snaphead is fitted to $N$ labeled images on frozen features, while SAM~3 runs zero-shot, so the comparison measures what a few labels add.
We report Cohen's $\kappa$, which is zero at chance whatever the class balance.

\begin{figure*}[t]
\centering
\includegraphics[width=\linewidth]{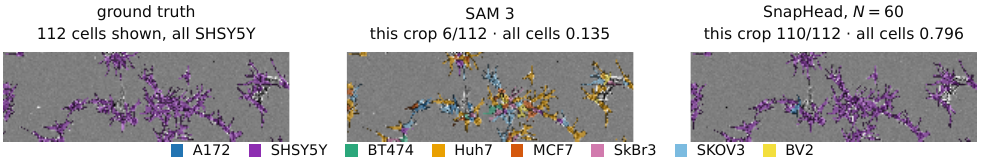}
\caption{Class assignment on one LIVECell image of SH-SY5Y cells (densest third, $112$ of $190$ cells). Every panel draws the same ground-truth masks; color shows the assigned line. Titles give the count right in this crop and the accuracy over all $35{,}458$ evaluation cells.}
\label{fig:qualitative}
\end{figure*}

\cref{tab:naming} gives the result.
On eight look-alike LIVECell cell lines, SAM~3 reaches $\kappa = 0.038$, and its accuracy of $0.135$ is below the $0.299$ of always guessing the commonest line.
The same protocol reaches $\kappa = 0.940$ on ten COCO animal species and $0.927$ on twenty rare LVIS classes.
\snaphead reaches $\kappa = 0.754$ on the cell lines at $N{=}60$, and $0.429$ from eight images, one per line.
On these two controls, where the cut lies in its vocabulary, SAM~3 stays ahead: it separates species but not cell lines.
SAM~3 scores each phrase independently, so its classes never compete for a mask.
Where its vocabulary already separates the classes, each phrase finds its own objects; on the cell lines, every phrase returns largely the same cells.
\snaphead fits one softmax over all classes (\cref{eq:head}), so the line between them is what it learns.
SAM~3's released image model accepts visual exemplars only from the image being segmented, so a user cannot demonstrate the cut from other images (\cref{sec:sup_sam3}); \cref{fig:qualitative} shows an example.

\paragraph{Words versus labels.}
To put look-alike classes in one image, we split LIVECell-A172 cells into area terciles and give the same split two sets of names (last two rows of \cref{tab:naming}).
SAM~3 reaches $\kappa = 0.125$ with ``small'', ``medium'' and ``large'', and chance ($-0.007$) with ``type A'', ``type B'' and ``type C''.
\snaphead learns from labels alone, so both namings give it the same score, and it rises with each labeled image instead: $0.068$ at $N{=}1$, $0.554$ at $N{=}5$ and $0.722$ at $N{=}60$.

\begin{table}[t]
\centering
\footnotesize
\setlength{\tabcolsep}{4pt}
\caption{Class assignment on shared ground-truth masks. Cohen's $\kappa$, zero at chance. SAM~3: zero-shot prompt oracle; ours is \snaphead at $N{=}60$. The last two rows split LIVECell-A172 into area terciles under two class-name sets. Bold: better of the two.}
\label{tab:naming}
\begin{tabular}{@{}lrrr@{}}
\toprule
Project & $K$ & SAM~3 & ours \\
\midrule
LIVECell, 8 cell lines & 8 & 0.038 & \textbf{0.754} \\
COCO animals & 10 & \textbf{0.940} & 0.392 \\
LVIS rare & 20 & \textbf{0.927} & 0.605 \\
\midrule
A172, small/medium/large & 3 & 0.125 & \textbf{0.722} \\
A172, type A/B/C & 3 & $-0.007$ & \textbf{0.722} \\
\bottomrule
\end{tabular}
\end{table}

\subsection{Effect of label granularity}
\label{sec:granularity}

If the class boundary is what changes between projects (\cref{sec:intro}), then on fixed objects a finer cut should hurt a generic model and cost \snaphead only more labels.
We test this on three datasets whose labels form a ladder: Stanford Cars by body type, make and model~\cite{krause2013cars}, FGVC-Aircraft by manufacturer, family and variant~\cite{maji2013aircraft}, and ShipRSImageNet by ship versus dock, category and hull type~\cite{zhang2021shiprsimagenet}.
Every method classifies the same mask, which SAM2 decodes from the dataset's box, on the same $546$, $300$ and $120$ evaluation images at every cut; SAM~3 again uses its prompt oracle, and \snaphead sees twenty labeled images per finest class.

\begin{figure}[t]
\centering
\includegraphics[width=\columnwidth]{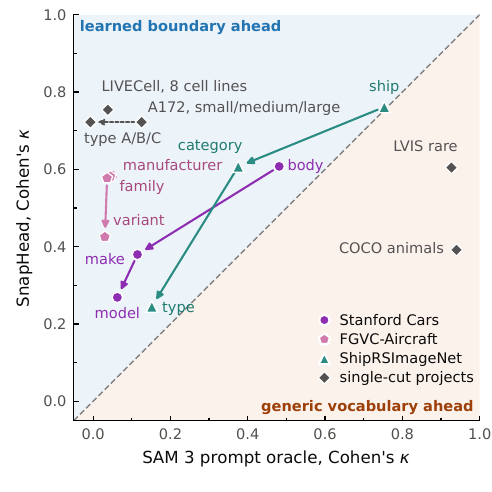}
\caption{Comparison with SAM~3 on the label ladders and class-assignment projects. Each point is one cut; above the diagonal, the learned boundary wins. Colored markers follow the three ladders, coarse to fine; \snaphead sees $20$ labeled images per finest class. Gray diamonds are the single-cut projects of \cref{tab:naming} at $N{=}60$. The dashed arrow renames the A172 size classes: SAM~3 moves, ours stays.}
\label{fig:granularity}
\end{figure}

\cref{fig:granularity} places every cut against SAM~3.
On cars, SAM~3 falls from $\kappa = 0.481$ at body type to $0.063$ at model, while \snaphead leads at every cut, from $0.608$ to $0.269$.
On aircraft, SAM~3 sits near chance at every cut ($\kappa \le 0.045$), below always answering Boeing at manufacturer (accuracy $0.100$ against $0.220$); \snaphead reaches $0.582$ there and $0.425$ at variant.
On ships, we match SAM~3 where the cut is ship versus dock ($0.761$ against $0.753$) and lead once it moves to category ($0.607$ against $0.375$) and hull type ($0.245$ against $0.152$).
On PanNuke, whose five nucleus types are in SAM~3's vocabulary, SAM~3 leads at the finest cut, as on COCO and LVIS (\cref{sec:sup_granularity}).

\subsection{Joint localization and class assignment}
\label{sec:proposer_res}
End to end, a method can miss objects before class assignment matters.
On LIVECell-A172, SAM2's automatic mask generator (AMG) recalls $0.167$ of the cells at IoU $0.5$, while a ground-truth box through the same frozen decoder recalls $0.880$, so the cells are lost at the proposal stage, not in the decoder (\cref{sec:sup_oracle}).
Our learned proposer (\cref{sec:proposer}) lifts recall to $0.406$ from one labeled image, $0.587$ from five and $0.644$ from twenty.
Instance F1 rises from $0.124 \pm 0.004$ with AMG to $0.432 \pm 0.046$ at $N{=}5$ and $0.481 \pm 0.003$ at $N{=}20$, and the drawn fraction $D$ falls from $0.932$ to $0.585$ and $0.504$.

The automatic switch (\cref{sec:proposals}) reads AMG's recall on the labeled images and adds the proposer only where that recall is low.
Over the 13 projects at $N{=}20$ it reaches mean F1 $0.239$, against $0.184$ for AMG alone and $0.234$ for always taking the union, within $0.003$ of an oracle that picks the better source per project on evaluation labels ($0.241$).

\paragraph{Comparison with baselines.}
\label{sec:master}
We group projects by how much of SAM~3's oracle F1 a blind choice of phrasing keeps: $0.623$--$0.750$ on the six natural projects, which we call robust, and $0.200$--$0.483$ on the other seven, which we call fragile (Mann-Whitney $p = 0.0012$, \cref{sec:sup_axis}).
Only the evaluation labels reveal which phrase works: on BBBC038 the class name ``nucleus'' scores $0.003$ and ``bright spots'' $0.742$, and a blind pick among the five expects $0.170$.
An independent microscopy benchmark reports the same pattern: SAM~3 does not recognize ``nucleus'', and words for the shape often beat the biological term~\cite{archit2026apg}.
\cref{tab:master} compares every method on the five microscopy projects at $N{=}20$, on the same images and seeds, with all three metrics of a method taken from one set of predictions.
The other two fragile projects are aerial, whose large tiles Mask~R-CNN cannot run on (recall in \cref{tab:oracle_full}).
With $20$ labeled images, the router (\cref{sec:router}) picks SAM~3's phrasing from them and reaches F1 $0.484$, above SAM~3's oracle ($0.441$; $+0.043$, 95\% CI $[0.031, 0.056]$, $p < 10^{-4}$, paired bootstrap over evaluation images, \cref{tab:bootstrap}), although the oracle tunes phrasing and threshold on the evaluation labels.
The labels send $0.467$ of the microscopy classes to SAM~3 and the rest to \snaphead, and each source alone scores lower: $0.420$ for SAM~3 at the phrasing chosen on $\mathcal{L}$ ($[0.051, 0.077]$ below the router) and $0.433$ for \snaphead ($[0.044, 0.059]$ below).
\snaphead alone, with no SAM~3 at all, comes within $0.008$ F1 of the oracle (95\% CI $[-0.024, 0.007]$) and leads SAM~3 with its class name by $0.214$ ($[0.195, 0.231]$, $p < 10^{-4}$).
It also outperforms SAM~3 with a blind phrasing, the pixel forest, PerSAM and Grounding DINO on all five projects.
An annotator without the answer key or labels gets the blind row, $0.106$, and more exemplars do not lift it: through the published few-shot recipe, SAM~3 gains at most $0.013$ F1 from $1$ to $20$ examples on any of six projects (\cref{tab:sam3fs}).
\snaphead improves with every label instead, from $0.290$ at $N{=}1$ to $0.433$ at $N{=}20$ and $0.449$ at $N{=}50$, and it matches few-shot Mask~R-CNN trained on the same labels ($+0.014$, $p = 0.50$).
Its drawn fraction, $0.578$, is $0.280$ below SAM~3 with its class name ($[0.262, 0.298]$), lower on all five projects, and $0.031$ below even SAM~3's oracle ($[0.012, 0.048]$, $p = 0.0004$).
The router's effort, which charges every deletion, is $0.694$, the lowest in \cref{tab:master} without $\ddagger$ and below SAM~3's oracle ($0.719$).
Compared like for like, \snaphead is the fastest in each trained group of \cref{tab:master} (\cref{sec:sup_adapt}).
On the six natural projects, where the cut already lies in SAM~3's vocabulary, \snaphead alone trails SAM~3 with class names ($0.114$ against $0.629$); there the labels send $0.911$ of the classes to SAM~3, and the router reaches $0.615$, close to SAM~3's $0.629$.
Over all $13$ projects the router reaches $0.503$, above SAM~3 with class names ($0.404$), SAM~3 at the phrasing chosen on $\mathcal{L}$ ($0.495$, $p = 0.003$) and \snaphead ($0.239$), and $0.008$ below the oracle ($0.511$) (\cref{tab:router_full}).

\begin{table}[t]
\centering
\footnotesize
\setlength{\tabcolsep}{1.5pt}
\caption{Five microscopy projects at $N{=}20$, mean over projects ($t$: median). $F1_{\mathrm{cl}}$, $F1_{\mathrm{nu}}$: $F1$ on LIVECell's eight cell lines and on BBBC038 nuclei. $D$: fraction drawn by hand. $E$: effort at $w_{\mathrm{check}}{=}0.1$, $w_{\mathrm{refine}}{=}0.5$; $[E]$: min--max of $E$ over the cited box, $w_{\mathrm{check}} \in [0.063,0.473]$~\cite{papadopoulos2016verify,kuznetsova2020open}, $w_{\mathrm{refine}} \in [0.268,0.5]$~\cite{oehme2026aramsam}. $t$: seconds to adapt, $^{c}$CPU, $^{g}$GPU; bold: fastest in group. Drawing every object by hand gives $D{=}E{=}1$. $\dagger$: oracle, phrasing and threshold tuned on evaluation labels. $\ddagger$: proposals from Cellpose, whose training data include LIVECell and BBBC038 (\cref{sec:sup_cellpose}). Bold: best without $\dagger$ or $\ddagger$.}
\label{tab:master}
\begin{tabular}{@{}lccccccr@{}}
\toprule
Method & $F1{\uparrow}$ & $F1_{\mathrm{cl}}{\uparrow}$ & $F1_{\mathrm{nu}}{\uparrow}$ & $D{\downarrow}$ & $E{\downarrow}$ & $[E]{\downarrow}$ & $t{\downarrow}$ \\
\midrule
SAM~3, name & 0.219 & 0.000 & 0.000 & 0.858 & 0.893 & 0.878--0.936 & 0 \\
SAM~3, blind & 0.106 & 0.016 & 0.170 & 0.911 & --- & --- & 0 \\
SAM~3$^\dagger$ & 0.441 & 0.078 & 0.742 & 0.609 & 0.719 & 0.672--0.904 & 0 \\
G-DINO$^\dagger$ & 0.090 & 0.000 & 0.359 & 0.940 & 0.956 & 0.950--0.991 & 0 \\
\midrule
pixel forest & 0.249 & 0.010 & 0.806 & 0.806 & 0.847 & 0.830--0.915 & $0.30^{c}$ \\
PerSAM & 0.253 & 0.097 & 0.704 & 0.741 & 0.815 & 0.785--0.967 & $0.07^{c}$ \\
ours, AMG & 0.285 & 0.129 & \textbf{0.842} & 0.762 & 0.814 & 0.792--0.901 & $\mathbf{0.03}^{c}$ \\
\midrule
Mask~R-CNN & 0.419 & 0.001 & 0.816 & 0.650 & 0.741 & 0.702--0.857 & $7.39^{g}$ \\
ours, gated & 0.433 & \textbf{0.216} & \textbf{0.842} & 0.578 & 0.736 & \textbf{0.668}--0.949 & $\mathbf{3.83}^{g}$ \\
\midrule
SAM~3, $\mathcal{L}$-phrase & 0.420 & 0.000 & 0.726 & 0.678 & 0.764 & --- & --- \\
ours, router & \textbf{0.484} & \textbf{0.216} & \textbf{0.842} & \textbf{0.572} & \textbf{0.694} & --- & --- \\
ours + Cellpose$^\ddagger$ & 0.544 & 0.338 & 0.845 & 0.495 & 0.655 & --- & --- \\
\bottomrule
\end{tabular}
\end{table}

\paragraph{Annotation effort.}
\label{sec:loop}
\label{sec:study}
\paper needs less effort than drawing by hand on all 13 development projects: at the default weights of \cref{eq:cost}, pooled $E = 0.807$ at $N{=}5$ and $0.784$ at $N{=}20$.
Inside the cited box, the minimum effort of \snaphead ($0.668$) is below even that of SAM~3's oracle ($0.672$), in the cheap-check region.
The effort dial (\cref{sec:effort}) bounds cost at manual: it holds on $231/231$ tuning curves and $221/231$ held-out splits ($95.7\%$); with it on, effort stays below manual over the whole cited box on all five microscopy projects (\cref{sec:sup_guarantee}).
Replaying the loop, $36$ rounds of five images on four projects, effort falls below manual from the second round on, every project and seed, staying below it over $0.988$--$1.000$ of the cited range (\cref{tab:loop}).
Most of what remains is drawing (pooled $D = 0.670$ of $E = 0.784$, \cref{fig:effort}).

\subsection{Generalization to held-out datasets}
\label{sec:transfer}

We ran the frozen configuration once, unchanged, at $N{=}20$ on the five RF100-VL datasets of \cref{sec:setup}, scoring at box IoU $0.5$ as RF100-VL has only boxes.
We count a win when our mean minus one standard deviation over seeds exceeds SAM~3's score.
By that rule we outperform SAM~3 with class names on four of the five datasets and its phrasing-and-threshold oracle on two (\cref{tab:transfer}); a paired bootstrap over evaluation images confirms both oracle wins ($p \le 0.0004$, \cref{tab:bootstrap}).
With SAM~3 as a second source, the frozen router reaches the best mean F1, $0.328$, against $0.308$ for \snaphead alone ($p = 0.0005$) and $0.306$ for SAM~3's oracle.
The clearest win is the finest cut in the draw, everdaynew (one product in different package sizes): we reach $0.294$ against $0.107$ for SAM~3 with class names and $0.145$ for its oracle.
Where SAM~3's oracle leads, the classes are everyday words (people and hats; bottle and can; humans), where a generic vocabulary should win.

\begin{table}[t]
\centering
\footnotesize
\setlength{\tabcolsep}{1.5pt}
\caption{Transfer to five randomly drawn RF100-VL datasets at $N{=}20$. Instance F1 at box IoU $0.5$, mean $\pm$ sd over three seeds; SAM~3 is deterministic. \emph{ours}: \snaphead. SAM~3: class names, default threshold. \emph{oracle}: best phrasing and threshold on evaluation labels. \emph{router}: \cref{sec:router}. SW: the automatic switch (\cref{sec:proposals}) is within seed noise of the better fixed proposer. Last row: mean; sd over seeds.}
\label{tab:transfer}
\begin{tabular}{@{}lccccc@{}}
\toprule
dataset (classes) & ours & router & SAM~3 & oracle & SW \\
\midrule
everdaynew {\scriptsize(pack sizes)} & $\mathbf{0.294}${\scriptsize$\pm$0.030} & $0.250${\scriptsize$\pm$0.042} & $0.107$ & $0.145$ & y \\
aerial-pool {\scriptsize(hats, people)} & $0.188${\scriptsize$\pm$0.009} & $0.317${\scriptsize$\pm$0.010} & $0.052$ & $\mathbf{0.354}$ & y \\
stomata {\scriptsize(open, closed)} & $\mathbf{0.260}${\scriptsize$\pm$0.012} & $0.239${\scriptsize$\pm$0.038} & $0.000$ & $0.154$ & n \\
floating-waste {\scriptsize(litter)} & $0.196${\scriptsize$\pm$0.027} & $0.233${\scriptsize$\pm$0.003} & $0.237$ & $\mathbf{0.253}$ & y \\
floods {\scriptsize(human)} & $0.602${\scriptsize$\pm$0.011} & $0.601${\scriptsize$\pm$0.011} & $0.578$ & $\mathbf{0.622}$ & y \\
\midrule
mean & $0.308${\scriptsize$\pm$0.014} & $\mathbf{0.328}${\scriptsize$\pm$0.017} & $0.195$ & $0.306$ & --- \\
\bottomrule
\end{tabular}
\end{table}
\section{Conclusion}
\label{sec:concl}

A generic segmenter's class boundary is fixed by its vocabulary; \paper{} learns each project's own boundary on the fly from its labels, on one frozen SAM2.1 pass per image.
Handed identical masks of eight look-alike cell lines, \snaphead reaches $\kappa{=}0.754$ against SAM~3's $0.038$, and holds $0.722$ when only the class names change, while SAM~3 moves from $0.125$ to $-0.007$.
With twenty labeled images \snaphead alone beats SAM~3's oracle on those cell lines ($0.216$ against $0.078$ instance F1), on nuclei ($0.842$ against $0.742$) and on two of five held-out RF100-VL projects.
Where the cut already lies in SAM~3's vocabulary, SAM~3's own class is better, so the two complement each other: keeping, per class, the boundary the labels favor beats SAM~3's oracle on microscopy ($0.484$ against $0.441$) and trails it by $0.008$ over all $13$ projects.
We build on SAM2.1 for its Apache~2.0 license; SAM~3 (SAM License~\cite{meta2025sam3repo}) stays optional.
A foundation model supplies where the objects are; the project's own labels decide which boundary to trust.
{
    \small
    \bibliographystyle{ieeenat_fullname}
    \bibliography{main}
}

\clearpage
\maketitlesupplementary
\setcounter{section}{0}
\setcounter{table}{0}
\setcounter{figure}{0}
\setcounter{equation}{0}
\renewcommand{\thesection}{\Alph{section}}
\renewcommand{\thetable}{A\arabic{table}}
\renewcommand{\thefigure}{A\arabic{figure}}
\renewcommand{\theequation}{A\arabic{equation}}
\renewcommand{\theHsection}{supp.\Alph{section}}
\renewcommand{\theHtable}{supp.\arabic{table}}
\renewcommand{\theHfigure}{supp.\arabic{figure}}
\renewcommand{\theHequation}{supp.\arabic{equation}}

In this supplementary material, we provide implementation details (\cref{sec:sup_impl}), dataset and protocol details (\cref{sec:sup_data}), the full comparison with SAM~3 (\cref{sec:sup_sam3}), additional results and ablations (\cref{sec:sup_results}), annotation effort (\cref{sec:sup_effort}), and runtime and memory cost (\cref{sec:sup_cost}).

\section{Implementation Details}
\label{sec:sup_impl}

\begin{table}[!h]
\centering
\footnotesize
\setlength{\tabcolsep}{4pt}
\caption{Default \paper configuration, used for every project. \emph{auto} entries are read from the labeled images by the rule shown; the rest are fixed.}
\label{tab:config}
\begin{tabular}{@{}p{2.0cm}p{5.9cm}@{}}
\toprule
Component & Setting \\
\midrule
Encoder & SAM2.1-Hiera-L, frozen, one forward pass per image \\
Proposer pool & SAM2's automatic mask generator (AMG), fixed knobs; plus a $97{,}603$-parameter proposer, architecture fixed, weights fit per project (\emph{auto}) \\
Pool switch & \emph{auto}: the union of AMG and the learned proposer once AMG's recall on $\mathcal{L}$ falls below $0.5$, AMG alone otherwise (\cref{sec:proposals}) \\
Descriptor & $x$: the mask-pooled SAM2 image embedding (256-d) joined with the box-prompt object pointer (256-d) \\
\snaphead & multinomial logistic regression, balanced class weights, standardized descriptors, explicit background class \\
Decision rule & one flat threshold, $0.5$, on \snaphead's class probability times SAM's mask-quality score, the same for every class and every project \\
Suppression & containment-aware: candidates linked at IoU $\ge 0.4$ or intersection-over-smaller-mask $\ge 0.8$ collapse to the one ranked highest by that same score \\
External boxes & SAM~3, Grounding DINO or YOLOE; optional, off by default (\cref{sec:absorption}) \\
\bottomrule
\end{tabular}
\end{table}

\paragraph{Proposals and descriptors.}
We run SAM2.1-Hiera-L frozen at production AMG density (64 points per side, one crop layer). The descriptor $x$ is the 256-d mean image embedding over the mask, $\ell_2$-normalized and joined with the 256-d box-prompt object pointer. We drop proposals whose best ground-truth IoU falls in an ambiguous band from training.

\paragraph{Learned proposer.}
AMG's point grid bottlenecks resolution on dense projects (\cref{sec:bottleneck}). LIVECell's stride-16 embedding spans a 30\,px cell; we reuse the stride-4 decoder features instead, for a stride of $2.75$\,px on a 704\,px frame.
Input: two high-resolution feature maps ($32{\times}256{\times}256$ at stride 4, $64{\times}128{\times}128$ at stride 8) and the image embedding ($256{\times}64{\times}64$), upsampled to $352{\times}256{\times}256$; cell $(g_x,g_y)$ maps to $((g_x{+}0.5)W/256, (g_y{+}0.5)H/256)$. Trunk: group normalization (11 groups, 352 channels), $1{\times}1$ conv to 64, two [$3{\times}3$ conv 64, group normalization (8 groups), ReLU] blocks, two $1{\times}1$ heads.

Center head: CornerNet's penalty-reduced focal loss ($\alpha{=}2$, $\beta{=}4$) on Gaussian-splatted centers (overlap $0.7$), plus $\ell_1$ on log size. Inference: sigmoid, $3{\times}3$ max-pool peak test, top-$K$ cut ($K{=}3\times$ median seed-image count, cap 512); SAM2 decodes, keeps the best mask, ranks by peak score times predicted IoU (NMS $0.7$). Training: 300 Adam steps, learning rate $3\mathrm{e}{-3}$, cosine decay, batch $\min(4,N)$.

Distance-map head: binary cross-entropy on foreground, $\ell_1$ on a per-instance distance map normalized to 1 at the deepest point; decoding seeds at $d \ge 0.5$ and grows by Euclidean influence zones. \cref{sec:sup_ablations} scores it against the center head.

The proposer has $97{,}603$ parameters and fits in under four seconds on cached features (\cref{tab:cost} gives per-image cost against AMG). Micro-SAM's decoder beats AMG on microscopy but needs a large annotated corpus~\cite{archit2025microsam}; our proposer fits from the images already labeled for the project.

\paragraph{Budgets and seeds.}
We use budgets $N \in \{1,2,5,10,20,50\}$, three seeds each, disjoint subsets nested in $N$. We read $K$ and the AMG-recall gate from labeled images. Below seed-set AMG recall $0.5$ the pool unions AMG and the learned proposer; at or above it, AMG runs alone. \cref{sec:sup_master} gives the band \paper uses by default.

\paragraph{Constants.}
We fix the proposer at $\kappa = 3$, $[K_{\min}, K_{\max}] = [16, 512]$; the switch uses $\tau_r = 0.5$. \snaphead uses scikit-learn's default weight ($C{=}1$), $\tau = 0.5$, $\tau_{\mathrm{nms}} = 0.4$; \texttt{accept} needs IoU $\tau_{\mathrm{acc}} = 0.75$, fixed across projects (\cref{sec:sup_cost}).

\paragraph{Metric choices.}
Our F1 at IoU $0.5$ follows CellSAM~\cite{marks2025cellsam}, stricter than mSA/COCO mask AP (IoU up to $0.95$, omitted here).
F1 pools all instances across the 13 projects rather than averaging over images.
Averaging over images instead raises F1 on 11 of 13, most on LIVECell-A172/micro-SAM ($+0.075$ each, $230.2$/$396.0$ instances per image). It falls for COCO vehicles ($-0.008$) and the eight-line mix ($-0.001$). We report pooled F1 throughout.

\section{Datasets and Evaluation Protocol}
\label{sec:sup_data}

\subsection{Coarse-to-fine ladders}
\label{sec:sup_granularity}

SAM~3 still finds the object: in all $72$ settings on cars and aircraft, every phrase localizes it at IoU $0.5$, and only the confidence score changes.
On a fourth ladder, PanNuke, SAM~3 leads at the finest cut, already in its vocabulary.

We give \cref{sec:granularity}'s per-cut numbers, two more ladders, and their construction and protocol.

\cref{tab:granularity} gives the per-cut numbers, labeled images fixed at $N$ per finest class and relabeled per cut.

\begin{table}[t]
\centering
\footnotesize
\setlength{\tabcolsep}{5pt}
\caption{Cohen's kappa per cut, Stanford Cars and FGVC-Aircraft. SAM~3: zero-shot prompt oracle; \snaphead, three seeds.}
\label{tab:granularity}
\begin{tabular}{llrrrr}
\toprule
 & & & & \multicolumn{2}{c}{ours, fixed} \\
\cmidrule(lr){5-6}
Dataset & Cut & $K$ & SAM~3 & $N{=}5$ & $N{=}20$ \\
\midrule
\multirow{3}{*}{Stanford Cars} & body type & 9 & 0.481 & 0.448 & 0.608 \\
 & make & 48 & 0.115 & 0.193 & 0.380 \\
 & model & 182 & 0.063 & 0.088 & 0.269 \\
\midrule
\multirow{3}{*}{FGVC-Aircraft} & manufacturer & 30 & 0.045 & 0.385 & 0.582 \\
 & family & 70 & 0.036 & 0.329 & 0.577 \\
 & variant & 100 & 0.030 & 0.185 & 0.425 \\
\bottomrule
\end{tabular}
\end{table}

\paragraph{Best-kappa rule.}
For SAM~3's cell in \cref{tab:granularity} we pick the phrasing and decision rule by kappa instead of by macro-F1 as in \cref{sec:naming}, which can only raise its score.

\paragraph{ShipRSImageNet.}
We cut ShipRSImageNet's ships three ways over 120 images: ship/dock ($K{=}2$, chance $0.386$), four categories ($K{=}4$, chance $0.224$), fifty hull types ($K{=}50$, chance $0.014$). SAM~3's oracle reaches $0.753$/$0.375$/$0.152$.
At $N \ge 20$ \snaphead leads every cut: $0.761$/$0.931$ ($N{=}20$/$60$) for ship/dock; $0.607$/$0.591$ for category; $0.245$ for hull type.

\paragraph{PanNuke.}
PanNuke's finest cut splits nuclei into five pathology types ($K{=}5$, chance $0.176$, 120 images), a vocabulary SAM~3 already has. Its oracle reaches kappa $0.724$ from the bare class names. \snaphead reaches $0.173$/$0.309$/$0.362$/$0.363$ at $N{=}1/5/20/60$, well short of SAM~3. Here the cut sits inside SAM~3's vocabulary, and it wins.

\paragraph{Dataset construction.}
Stanford Cars pairs each Hugging Face mirror image\footnote{\url{https://huggingface.co/datasets/tanganke/stanford_cars}} to the devkit's box by assumed filename order. Shifting it by one image puts $44.9\%$ of boxes outside their image, a random pairing $46.2\%$, versus none assumed. We drop classes with no body word (14 of 196), leaving $182$ models and $48$ makes over $546$ images.
FGVC-Aircraft's box file is 1-based inclusive; $130$ of its $10{,}000$ boxes start at pixel $0$ and are clipped, $65$ here.
We score every arm on SAM2's decoded mask, at median IoU $0.954$/$0.961$ (cars/aircraft).

\paragraph{The iSAID-vehicles subsample.}
iSAID-vehicles has an oracle estimated on a subsample. Its tiles carry up to $7{,}374$ objects, so we cap ground truth at $1000$ per image, capping $12$ of its $179$ images and leaving $31{,}515$ of $53{,}713$ instances prompted, seeded and unbiased.

\paragraph{Aerial projects.}
We append both iSAID projects to the grid under \paper's gated system. With our descriptor both sit at a resolution floor: most iSAID-vehicles instances are under $8$\,px after SAM2's $1024$-square resize, below any method's workable size (\cref{sec:sup_oracle} prices the same floor on iSAID-buildings). We skipped few-shot Mask~R-CNN on both: torchvision caps the long side at $1333$\,px, so a $13{,}383$\,px tile loses its sub-$8$\,px instances.

\paragraph{SAM~3 oracle's two-collection rule.}
For the SAM~3 oracle row in \cref{tab:master_full}, we take the better of two prompt collections per project, both fixed in advance. It switches to the later matched-$K$ set of \cref{sec:sup_axis} only past run-to-run noise ($0.005$ F1). Only livecell-8type switches: correct cell-line names score $0.000$, an appearance description $0.078$.

\section{Comparison with SAM~3}
\label{sec:sup_sam3}

\subsection{Class assignment}

\begin{figure*}[t]
\centering
\includegraphics[width=\linewidth]{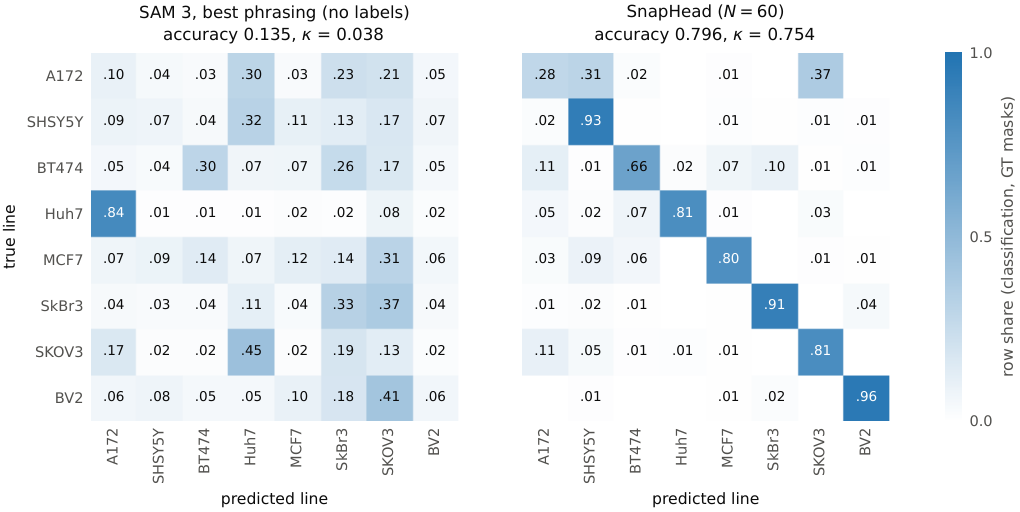}
\caption{Class assignment on eight LIVECell cell lines, SAM~3 against \snaphead. Masks fixed; $n{=}35{,}458$ instances, row-normalized, cells below $0.005$ blank.}
\label{fig:confusion}
\end{figure*}

\cref{fig:confusion} compares class assignment on the eight LIVECell cell lines. SAM~3 runs zero-shot with the phrasing that scores best on the evaluation labels; \snaphead is fitted to $N{=}60$ labeled images per line. We score instance classification on fixed masks, rather than the segmentation F1 of the main paper.

\paragraph{Reading the matrices.}
SAM~3's matrix on the eight cell lines is diffuse, mean diagonal $0.141$, its largest error Huh7 read as A172 at $0.84$. Per-line recognition ranges from SkBr3 at $0.33$ to Huh7 at $0.01$. \snaphead's matrix is a diagonal, mean $0.770$, peaking at BV2 with $0.96$. The same code and prompts on ten COCO animal species ($n{=}246$) give SAM~3 a near-perfect diagonal, mean $0.957$, with largest errors of $0.08$ and $0.05$.

\paragraph{Qualitative image.}
We show one SH-SY5Y image with $190$ instances in Fig.~\ref{fig:qualitative}, picked by the lower median of per-image instance count over the $120$-image evaluation split; its panels crop the image to its densest third, $112$ of the $190$ instances, at about the same SAM~3 error rate ($\sim 0.95$) as the full image.

\paragraph{Tercile split.}
The words-versus-labels test of \cref{sec:naming} splits the $27{,}623$ LIVECell-A172 instances into area terciles. All three classes are found in $118$ of the $120$ evaluation images. At $N{=}60$ \snaphead scores macro-F1 $0.816$ under both namings.

\paragraph{Text-encoder check.}
The text encoder keeps ``A172 cell'' and ``SHSY5Y cell'' apart. Raw phrase cosines are about $0.99$ for every prompt set, yet the eight line names are mutually distinguishable after mean-centering. We therefore attribute the failure to vision-language alignment.

\paragraph{Architectural limit and composition.}
At most $200$ masks are emitted per prompt by SAM~3, one per object query. On a project averaging $295$ instances per image, recall caps at $0.520$ regardless of phrasing. Restricted to the instances a phrase actually returns, SAM~3 scores $0.150$ against a $0.121$ chance level. The per-instance best-IoU spread across the eight phrases is $0.08$ to $0.13$. Combining models beats either alone: a generic ``cell'' prompt with \snaphead assigning the class reaches $0.766$ on the eight lines, against $0.752$ for it on ground-truth boxes, both at $N{=}20$.

\paragraph{Pairwise separation.}
We score each of the $28$ cell-line pairs as a two-way decision, so $0.5$ is chance. \snaphead's accuracy ranges from $0.544$ to $0.998$ (median $0.971$). SAM~3's ranges from $0.497$ to $0.657$ (median $0.547$), with $25$ of the $28$ pairs below $0.6$; it leads on all $28$.

\subsection{Prompts and exemplars}

\paragraph{Prompts.}
We add the description phrasing because SAM~3's own prompt study reports $0.085$ mSA for ``nucleus'' and $0.509$ for ``bright spots'' on the Data Science Bowl. We fix all $530$ distinct strings in advance. The same five phrasings per class are used on every project (\cref{sec:sup_axis}). Grounding DINO~\cite{liu2024groundingdino} and YOLOE~\cite{wang2025yoloe} run the same phrasings; SAM~3 beats both on 12 of the 13 projects (\cref{tab:detectors}), and on the eight cell lines all three score near zero, so we report SAM~3.

\begin{table}[t]
\centering
\footnotesize
\setlength{\tabcolsep}{4pt}
\caption{Oracle F1 for SAM~3, Grounding DINO and YOLOE under one protocol: $120$ evaluation images, the same five phrasings per class, each model's own oracle over seven thresholds. Bold: best per row. SAM~3 uses this shared protocol without the matched-$K$ prompt set of \cref{sec:sup_axis}, so livecell-8type reads $0.000$ here against $0.078$ in \cref{tab:master}. That row is unbolded: all three round to $0.000$ (Grounding DINO $0.0002$).}
\label{tab:detectors}
\begin{tabular}{lrrr}
\toprule
Project & SAM~3 & Grounding DINO & YOLOE \\
\midrule
coco-animals & \textbf{0.914} & 0.806 & 0.780 \\
voc-20 & \textbf{0.829} & 0.777 & 0.735 \\
coco-vehicles & \textbf{0.766} & 0.379 & 0.652 \\
coco-kitchen & \textbf{0.636} & 0.279 & 0.476 \\
lvis-rare20 & \textbf{0.439} & 0.116 & 0.207 \\
ade20k-indoor & \textbf{0.379} & 0.218 & 0.305 \\
\midrule
bbbc038-nuclei & \textbf{0.742} & 0.359 & 0.000 \\
livecell-A172 & \textbf{0.586} & 0.057 & 0.000 \\
livecell-microsam & \textbf{0.424} & 0.022 & 0.000 \\
livecell-SHSY5Y & \textbf{0.375} & 0.010 & 0.000 \\
isaid-vehicles & \textbf{0.304} & 0.067 & 0.002 \\
isaid-buildings & \textbf{0.172} & 0.038 & 0.001 \\
livecell-8type & 0.000 & 0.000 & 0.000 \\
\bottomrule
\end{tabular}
\end{table}

\paragraph{Exemplars.}
The released image API has no image-level exemplar slot. \cref{tab:sam3fs} gives the full few-shot grid. Under the published spatial-canvas workaround~\cite{tsai2026fewshot}, no project improves by more than $0.001$ from $N{=}5$ to $N{=}20$, and five of six fall.

\subsection{Few-shot grid}

\paragraph{Few-shot recipe.}
\cref{tab:sam3fs} uses the published spatial-canvas recipe~\cite{tsai2026fewshot}: up to three support panels and four canvases per class, one exemplar box per panel. Detections are unioned over canvases and NMS'd at IoU $0.7$. The $N{=}1$ canvas holds two panels against four for $N{\ge}5$, so part of the $1{\to}5$ drop is panel resolution.

\begin{table}[t]
\centering
\footnotesize
\caption{SAM~3 few-shot by the spatial-canvas recipe: six projects, $N \in \{1,5,20\}$, three seeds, a $60$-image evaluation prefix; full coverage in \cref{sec:sup_master}. \emph{text}: the matched-$K$ zero-shot oracle of \cref{tab:axis}. Bold: best per project.}
\label{tab:sam3fs}
\begin{tabular}{lcccr}
\toprule
Project & $N{=}1$ & $N{=}5$ & $N{=}20$ & text \\
\midrule
coco-animals & 0.884 & 0.896 & 0.893 & \textbf{0.916} \\
coco-vehicles & 0.726 & 0.715 & 0.711 & \textbf{0.765} \\
lvis-rare20 & 0.520 & 0.533 & \textbf{0.533} & 0.439 \\
bbbc038-nuclei & 0.756 & 0.676 & 0.634 & \textbf{0.742} \\
livecell-A172 & 0.256 & 0.219 & 0.219 & \textbf{0.586} \\
livecell-8type & 0.092 & \textbf{0.097} & 0.083 & 0.078 \\
\bottomrule
\end{tabular}
\end{table}

\subsection{Fine-tuning}

SAM~3's own $10$-shot fine-tuning raises its RF100-VL box AP from $15.2$ to $36.5$~\cite{carion2026sam3}. That setting fine-tunes the whole model per dataset. \snaphead instead fits on a frozen encoder inside the annotation loop.

\subsection{Phrasing sensitivity per project}
\label{sec:sup_axis}
\label{sec:fragility}

We prompt SAM~3 with five phrasings per class on all $13$ projects. An oracle picking the best phrasing and threshold on evaluation labels beats a blind pick on every project, by $0.062$ to $0.573$ F1.
SAM~3's worst of five phrasings never falls below $0.079$ on a natural project and never rises above $0.009$ on a scientific one. Six of seven scientific projects have a phrasing scoring exactly zero; no natural project does. The retained fraction, $\mathrm{exp}/\mathrm{oracle}$, is $0.62$ to $0.75$ on natural projects and $0.20$ to $0.48$ on scientific ones (Mann-Whitney $p{=}0.0012$, six against seven projects).
On BBBC038, ``nucleus'' scores $0.003$ against $0.742$ for ``bright spots.'' On LIVECell-A172, ``cell'' scores $0.586$, and one true adjective drops it to $0.000$.

\cref{tab:axis} gives the per-project detail, all of it SAM~3.

\paragraph{Domain dependence.}
On scientific projects, the per-slot means show where phrasing fails. Only two slots fire there: the name at $0.266$ and the appearance phrase at $0.177$; synonym ($0.054$), hypernym ($0.006$) and qualified ($0.031$) score near zero. The appearance slot is weakest on natural images and the rescue on nuclei and the eight cell lines, so the two domains want opposite phrasing.
The class name is the best of the five phrasings on nine of the $13$ projects. It fails hardest on the rest.

\begin{table}[t]
\centering
\footnotesize
\setlength{\tabcolsep}{3.5pt}
\caption{Fragility axis, all 13 projects. SAM~3 under five phrasings per class ($K{=}5$), each at its best threshold; 120 evaluation images, 119 for the two aerial projects. \emph{oracle}: best phrasing on evaluation labels, in bold. \emph{exp}: expected F1 of a blind pick. $\Delta$: their difference. \emph{worst}: weakest phrasing. \emph{ret.}: $\mathrm{exp}/\mathrm{oracle}$, kept by a blind pick.}
\label{tab:axis}
\begin{tabular}{lrrrrrr}
\toprule
Project & cls & oracle & exp & $\Delta$ & worst & ret. \\
\midrule
coco-animals & 10 & \textbf{0.916} & 0.628 & 0.288 & 0.353 & 0.69 \\
voc-20 & 20 & \textbf{0.822} & 0.616 & 0.205 & 0.274 & 0.75 \\
coco-vehicles & 5 & \textbf{0.765} & 0.477 & 0.288 & 0.082 & 0.62 \\
coco-kitchen & 11 & \textbf{0.636} & 0.421 & 0.215 & 0.194 & 0.66 \\
lvis-rare20 & 20 & \textbf{0.439} & 0.326 & 0.113 & 0.199 & 0.74 \\
ade20k-indoor & 22 & \textbf{0.375} & 0.241 & 0.133 & 0.079 & 0.64 \\
\midrule
bbbc038-nuclei & 1 & \textbf{0.742} & 0.170 & 0.573 & 0.000 & 0.23 \\
livecell-A172 & 1 & \textbf{0.586} & 0.117 & 0.469 & 0.000 & 0.20 \\
livecell-microsam & 1 & \textbf{0.424} & 0.125 & 0.299 & 0.000 & 0.30 \\
livecell-SHSY5Y & 1 & \textbf{0.375} & 0.104 & 0.271 & 0.000 & 0.28 \\
isaid-vehicles & 2 & \textbf{0.304} & 0.147 & 0.157 & 0.000 & 0.48 \\
isaid-buildings & 4 & \textbf{0.172} & 0.069 & 0.103 & 0.009 & 0.40 \\
livecell-8type & 8 & 0.078 & 0.016 & 0.062 & 0.000 & 0.20 \\
\bottomrule
\end{tabular}
\end{table}

\section{Additional Results and Ablations}
\label{sec:sup_results}

\subsection{Learned proposer}
\label{sec:sup_proposer}

\begin{figure*}[t]
\centering
\includegraphics[width=\linewidth]{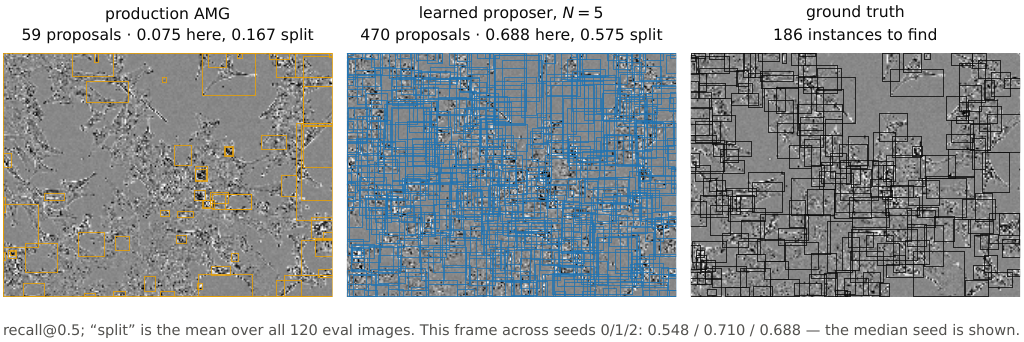}
\caption{Proposals on one LIVECell-A172 image: production AMG, the learned proposer ($N{=}5$, top-$K{=}512$ boxes), and ground truth, chosen by \cref{fig:qualitative}'s rule (\cref{sec:sup_sam3}). Panels give frame recall.}
\label{fig:proposals}
\end{figure*}

\cref{tab:proposer} gives proposal recall at $N{=}5$, and \cref{fig:proposals} shows the proposals on one image.

\begin{table}[t]
\centering
\footnotesize
\setlength{\tabcolsep}{4pt}
\caption{Proposal recall at IoU $0.5$, $N{=}5$, mean over 3 seeds and 120 images. \emph{LAP}: AMG knobs plus a stride-16 box scorer. \emph{oracle}: decoder ceiling. Bold: best below the oracle.}
\label{tab:proposer}
\begin{tabular}{lrrrr}
\toprule
Project & AMG & LAP\,$\cup$\,AMG & ours\,$\cup$\,AMG & oracle \\
\midrule
livecell-A172 & 0.167 & 0.295 & \textbf{0.587} & 0.880 \\
livecell-8type & 0.259 & 0.309 & \textbf{0.489} & 0.816 \\
livecell-microsam & 0.248 & 0.291 & \textbf{0.486} & 0.809 \\
livecell-SHSY5Y & 0.108 & 0.155 & \textbf{0.320} & 0.581 \\
\midrule
bbbc038-nuclei & 0.876 & 0.891 & 0.893 & 0.972 \\
ade20k-indoor & 0.506 & \textbf{0.566} & 0.529 & 0.715 \\
\bottomrule
\end{tabular}
\end{table}

\paragraph{Proposer image.}
We pick the image with \cref{fig:qualitative}'s rule (\cref{sec:sup_sam3}): seed $2$, the median learned-proposer recall across seeds ($0.548$, $0.710$, $0.688$). The learned panel draws $470$ boxes against $186$ ground-truth cells, reaching recall $0.688$ against a split mean of $0.575$. AMG reaches $0.075$ against a split mean of $0.167$.

\paragraph{Fine-tuned detector, per project.}
Across the 11 core projects, we beat few-shot Mask~R-CNN on instance F1 on $9$, $10$ and $9$ of $11$ at $N{=}1$, $5$ and $20$. We also beat it on drawn fraction, $11$, $11$ and $9$ of the same budgets. Pooled over three seeds, F1 is $0.141/0.202/0.259$ against $0.054/0.162/0.215$ at $N{=}1,5,20$.
On LIVECell-A172 ($230$ instances per image), the detector leads F1 at $N{=}2,5,20,50$ ($0.370/0.510/0.620/0.600$ against $0.307/0.432/0.481/0.484$); we lead at $N{=}1,10$ ($0.249/0.479$ against $0.059/0.365$).
On BBBC038 we lead on F1 at every budget: $0.819$, $0.823$, $0.825$, $0.825$, $0.842$ and $0.843$ against $0.413$, $0.614$, $0.725$, $0.777$, $0.816$ and $0.834$ at $N{=}1,2,5,10,20,50$.
A score of exactly $0.000$ is recorded for the detector on 3 of the 11 core projects at $N{=}1$ (ade20k-indoor, lvis-rare20, voc-20). Our proposer (\cref{sec:sup_impl}) uses a fraction of its $44$M parameters and $40$ steps.

\paragraph{AMG-strong projects.}
Adding the proposer where AMG already recalls well is mostly neutral. At $N{=}5$ on BBBC038, F1 is $0.825 \pm 0.003$/$0.780 \pm 0.035$/$0.573 \pm 0.058$ for AMG alone/the union/proposals alone. Drawn fraction is $0.199 \pm 0.013$/$0.203 \pm 0.013$/$0.517 \pm 0.106$. On ADE20K-indoor, proposals alone reach only $0.005 \pm 0.002$ and AMG wins outright.
AMG's recall on the \emph{seed} images is measurable from labels alone, so the switch adds no cost.

\subsection{Oracle decomposition}
\label{sec:sup_oracle}

\paragraph{Frozen-decoder ceiling.}
\label{sec:bottleneck}
\cref{tab:oracle_full} aggregates recall per image, unlike \cref{tab:master}'s instance-pooled F1, so the two scales differ. The gap matters where object sizes are skewed. iSAID-vehicles' box oracle is $0.534$ image-macro but, once instances are pooled, only $0.228$, since $82\%$ of its instances sit under $8$\,px at model input and crowd a few dense tiles.

Error splits into ``the proposer never found it'' and ``\snaphead got it wrong.'' Natural projects and nuclei are classification-bound. AMG recalls $0.57$ to $0.91$ of ground truth, and it is the limit. LIVECell is the reverse. AMG recalls only $0.058$ to $0.181$, and it is correct on $0.72$ to $0.94$ of what it sees. A ground-truth box recovers $0.581$ to $0.880$ there. Centroid points do not help either ($0.13$ to $0.26$).
An independent 2026 benchmark measures unmodified SAM's AMG at $0.075$ mSA on LIVECell and $0.331$ on the Data Science Bowl~\cite{archit2026apg}.

\begin{table}[t]
\centering
\footnotesize
\setlength{\tabcolsep}{3.5pt}
\caption{Recall and \snaphead accuracy per project at IoU $0.5$, image-macro; aerial rows cover 179 images (iSAID-vehicles box oracle: subsample, below). \emph{box}/\emph{point}: oracle recall from a ground-truth box or centroid. \emph{gap}: box oracle minus AMG. \emph{\snaphead}: accuracy on instances AMG hits.}
\label{tab:oracle_full}
\begin{tabular}{lrrrrrr}
\toprule
Project & $n_{\mathrm{GT}}$ & AMG & box & point & gap & \snaphead \\
\midrule
coco-animals & 452 & 0.912 & 0.990 & 0.902 & 0.078 & 0.433 \\
bbbc038-nuclei & 6951 & 0.851 & 0.972 & 0.889 & 0.121 & 0.954 \\
voc-20 & 396 & 0.760 & 0.937 & 0.740 & 0.177 & 0.357 \\
coco-kitchen & 667 & 0.748 & 0.883 & 0.709 & 0.135 & 0.181 \\
coco-vehicles & 687 & 0.737 & 0.958 & 0.710 & 0.220 & 0.523 \\
lvis-rare20 & 1019 & 0.657 & 0.948 & 0.689 & 0.291 & 0.360 \\
ade20k-indoor & 681 & 0.565 & 0.715 & 0.448 & 0.149 & 0.429 \\
\midrule
livecell-8type & 53820 & 0.181 & 0.816 & 0.261 & 0.635 & 0.725 \\
livecell-microsam & 68193 & 0.160 & 0.809 & 0.263 & 0.649 & 0.932 \\
livecell-A172 & 38500 & 0.096 & 0.880 & 0.210 & 0.784 & 0.944 \\
livecell-SHSY5Y & 66907 & 0.058 & 0.581 & 0.128 & 0.522 & 0.911 \\
\midrule
isaid-buildings & 4232 & 0.380 & 0.594 & 0.440 & 0.214 & 0.351 \\
isaid-vehicles & 53713 & 0.292 & 0.534 & 0.427 & 0.242 & 0.695 \\
\bottomrule
\end{tabular}
\end{table}

\subsection{All methods on one split}
\label{sec:sup_master}

\cref{tab:master_full} is \cref{tab:master} with the robust bin restored. \cref{sec:master} states the robust-bin result: SAM~3 reaches $0.661$ with its oracle phrasing and $0.629$ with the class name, against our $0.114$.

\begin{table}[t]
\centering
\footnotesize
\setlength{\tabcolsep}{3pt}
\caption{Every method, both bins, $N{=}20$. 11 core projects; the two aerial projects are analyzed separately, by recall, in \cref{tab:oracle_full}. \emph{drawn}: fraction of objects still drawn by hand. \emph{corr}: corrective actions per object. \emph{manual} has no F1 by construction. Every \emph{ours} row uses the same descriptor $x$. Bold: best F1 and drawn fraction per bin.}
\label{tab:master_full}
\begin{tabular}{lrrrrr}
\toprule
 & \multicolumn{3}{c}{microscopy (5)} & \multicolumn{2}{c}{robust (6)} \\
\cmidrule(lr){2-4}\cmidrule(lr){5-6}
Method & F1 & drawn & corr & F1 & drawn \\
\midrule
manual & --- & 1.000 & 1.000 & --- & 1.000 \\
\midrule
SAM~3, class name & 0.219 & 0.858 & 0.924 & 0.629 & 0.311 \\
SAM~3, blind phrasing & 0.106 & 0.911 & 0.998 & 0.452 & 0.517 \\
SAM~3, oracle & \textbf{0.441} & 0.609 & 1.002 & \textbf{0.661} & 0.338 \\
G-DINO, oracle & 0.090 & 0.940 & 1.001 & 0.429 & 0.448 \\
Mask~R-CNN, few-shot & 0.419 & 0.650 & 0.850 & 0.045 & 0.900 \\
\midrule
pixel forest & 0.249 & 0.806 & 0.885 & 0.017 & 0.720 \\
pixel forest, argmax & 0.267 & 0.773 & 0.893 & 0.001 & 0.999 \\
PerSAM reduction & 0.253 & 0.741 & 1.041 & 0.058 & \textbf{0.273} \\
\midrule
ours & 0.285 & 0.762 & 0.884 & 0.114 & 0.661 \\
ours, $\cup$ zero-shot & 0.419 & 0.623 & 1.013 & 0.128 & 0.582 \\
ours, effort dial & 0.246 & 0.796 & 0.873 & 0.058 & 0.846 \\
ours, gated & 0.433 & \textbf{0.578} & 1.153 & 0.114 & 0.661 \\
\bottomrule
\end{tabular}
\end{table}

\paragraph{Significance.}
Every significant result in \cref{tab:master_full} lands at exactly $p = 0.03125$, the smallest value a one-sided Wilcoxon signed-rank test returns over five projects.
A paired bootstrap over evaluation images has no such floor (\cref{sec:sup_bootstrap}).

\paragraph{Effort columns.}
The PerSAM reduction reaches the best drawn fraction in \cref{tab:master_full} ($0.273$, robust bin), but scores only $0.058$ F1 and $21.0$ corrective actions per object. Almost everything is predicted and the annotator deletes the rest, so we report drawn fraction and corrections per object together.

\subsection{Exemplar baselines at their published settings}
\label{sec:sup_exemplar}
We also run PerSAM, PerSAM-F~\cite{zhang2024personalize} and Matcher~\cite{liu2024matcher} from the authors' code at their published operating points, on the five microscopy projects with the same $20$ labeled images, seeds and scorer as \cref{tab:master} (\cref{tab:exemplar}).
Matcher uses up to $20$ support images per class.
Neither method handles several classes, so each class runs on its own and overlapping masks across classes are merged at IoU $0.5$.
At these settings no method passes $0.045$ mean F1: the published loops stop after a few objects, while a LIVECell image holds hundreds of cells.
Lifting the object caps raises PerSAM to $0.193$, still below the reduction we report in \cref{tab:master} ($0.253$) and well below \snaphead ($0.433$).

\begin{table*}[t]
\centering
\footnotesize
\setlength{\tabcolsep}{2.2pt}
\caption{Exemplar baselines at their published settings, five microscopy projects, $N{=}20$, instance F1 (mean over three seeds). $^\ast$: object cap lifted.}
\label{tab:exemplar}
\begin{tabular}{@{}lccccccc@{}}
\toprule
method & 8 lines & A172 & SH-SY5Y & micro-SAM & nuclei & mean & $D$ \\
\midrule
Matcher & 0.003 & 0.001 & 0.001 & 0.002 & 0.063 & 0.014 & 0.990 \\
PerSAM & 0.008 & 0.008 & 0.002 & 0.006 & 0.203 & 0.045 & 0.969 \\
PerSAM-F & 0.008 & 0.009 & 0.003 & 0.005 & 0.171 & 0.039 & 0.972 \\
Matcher$^\ast$ & 0.007 & 0.007 & 0.002 & 0.006 & 0.290 & 0.062 & 0.948 \\
PerSAM$^\ast$ & 0.037 & 0.063 & 0.014 & 0.087 & 0.766 & 0.193 & 0.804 \\
PerSAM-F$^\ast$ & 0.029 & 0.065 & 0.015 & 0.062 & 0.585 & 0.151 & 0.837 \\
\midrule
PerSAM reduction & 0.097 & 0.128 & 0.105 & 0.231 & 0.704 & 0.253 & 0.741 \\
ours, gated & \textbf{0.216} & \textbf{0.481} & \textbf{0.236} & \textbf{0.390} & \textbf{0.842} & \textbf{0.433} & \textbf{0.578} \\
\bottomrule
\end{tabular}
\end{table*}

\paragraph{Pixel forest.}
A random forest on the same frozen SAM~2 features ties us on BBBC038 (argmax variant, $0.834$ against $0.842$ at $N{=}20$). It loses across the bin ($0.267$ against $0.433$), close only where the proposer does no work.

\paragraph{Mask~R-CNN across budgets.}
We compare against few-shot Mask~R-CNN project by project in \cref{sec:sup_proposer}; it leads on two dense single-class projects, LIVECell-A172 and the LIVECell split defined by micro-SAM.

\paragraph{Gate threshold.}
Our system switches below a seed-set AMG recall of $0.35$. \cref{tab:master} and \cref{tab:master_full} use $0.5$ instead, disagreeing on four of $198$ seed-level cells.
Seed-set recall is close to bimodal. $0.024$--$0.422$ on the four LIVECell projects, $0.389$--$1.000$ on five of six natural ones, except BBBC038 ($0.787$--$1.000$) and LVIS-rare, spanning both modes ($0.153$--$1.000$). Recomputed at $0.35$, no bin mean moves more than $0.005$ F1 or $0.004$ drawn, unchanged at $N \ge 5$.
Any threshold from $0.259$ to $0.506$ matches $0.5$'s switch choice on the 13 projects. Helped projects read at or below $0.259$, unhelped ones at or above $0.506$; our default, $0.35$, sits inside.

\paragraph{Switch versus oracle choice.}
Two fixed policies bracket the switch on the five microscopy projects at $N{=}20$: AMG alone reaches $0.285$ F1, the learned-proposer union $0.429$, and the oracle choice $0.433$. Using only seed-set AMG recall, our switch matches the oracle choice on all five at $N{=}20$ and four of five at $N{=}1$. Livecell-8type costs it $0.001$ F1 there.

\subsection{Router per project}
\label{sec:sup_router}
\Cref{tab:router_full} lists every project at $N{=}20$ for the router of \cref{sec:router} and its inputs.
\emph{SAM~3, labeled} picks its phrasing set and threshold on the $20$ labeled images, the evidence the router also uses; it is the operating point an annotator with labels reaches without the answer key.
The router sends $0.911$ of natural classes to SAM~3, $0.792$ of aerial ones and $0.467$ of microscopy ones ($0.895$, $0.667$ and $0.333$ when SAM~3 keeps its class name).
On microscopy it reaches $0.484$ F1, above both of its inputs (\snaphead $0.433$, SAM~3 labeled $0.420$) and above SAM~3's oracle ($0.441$); over all 13 projects it reaches $0.503$, against $0.511$ for the oracle.
With $20$ labeled images, a class with few instances can be routed on noise: on three many-class natural projects, LVIS rare, VOC-20 and COCO kitchen, the router scores below SAM~3 with its class name, and each further labeled image adds instances to that per-class estimate.

\begin{table*}[t]
\centering
\footnotesize
\setlength{\tabcolsep}{2.5pt}
\caption{Router per project, $N{=}20$, mean over three seeds. \emph{\snaphead}: our head with the gated proposer. \emph{name}: SAM~3 with the class name at threshold $0.5$. \emph{labeled}: SAM~3 with phrasing and threshold chosen on the labeled images. \emph{r.\ name}: per-class choice between \snaphead and SAM~3 with its class name. \emph{router}: per-class choice that also picks SAM~3's phrasing on the labeled images (\cref{sec:router}). $\dagger$: oracle, phrasing and threshold chosen on evaluation labels. $D$, $E$ as in \cref{tab:master}. Bold: best F1 without $\dagger$. $^{*}$Two seeds.}
\label{tab:router_full}
\begin{tabular}{@{}l cccccc cccc cccc@{}}
\toprule
 & \multicolumn{6}{c}{F1$\uparrow$} & \multicolumn{4}{c}{$D\downarrow$} & \multicolumn{4}{c}{$E\downarrow$} \\
\cmidrule(lr){2-7}\cmidrule(lr){8-11}\cmidrule(lr){12-15}
Project & \snaphead & name & labeled & r.\ name & router & SAM~3$^\dagger$ & \snaphead & name & labeled & router & \snaphead & name & labeled & router \\
\midrule
ADE20K indoor & 0.077 & 0.313 & \textbf{0.362} & 0.314 & 0.344 & 0.379 & 0.763 & 0.456 & 0.467 & 0.493 & 0.899 & 0.947 & 0.837 & 0.835 \\
COCO animals & 0.177 & 0.881 & \textbf{0.898} & 0.881 & \textbf{0.898} & 0.914 & 0.411 & 0.073 & 0.106 & 0.106 & 0.506 & 0.254 & 0.267 & 0.267 \\
COCO kitchen & 0.081 & 0.620 & \textbf{0.629} & 0.587 & 0.596 & 0.636 & 0.836 & 0.325 & 0.404 & 0.470 & 0.913 & 0.514 & 0.557 & 0.604 \\
COCO vehicles & 0.216 & \textbf{0.766} & 0.745 & \textbf{0.766} & 0.745 & 0.766 & 0.482 & 0.198 & 0.265 & 0.265 & 0.621 & 0.371 & 0.415 & 0.415 \\
LVIS rare & 0.045 & \textbf{0.408} & 0.406 & 0.327 & 0.335 & 0.439 & 0.904 & 0.682 & 0.598 & 0.669 & 0.976 & 0.739 & 0.684 & 0.754 \\
VOC-20 & 0.089 & 0.785 & \textbf{0.819} & 0.713 & 0.771 & 0.829 & 0.572 & 0.134 & 0.152 & 0.220 & 0.657 & 0.334 & 0.337 & 0.378 \\
\emph{mean, natural} & 0.114 & 0.629 & \textbf{0.643} & 0.598 & 0.615 & 0.661 & 0.661 & 0.311 & 0.332 & 0.370 & 0.762 & 0.527 & 0.516 & 0.542 \\
\midrule
iSAID buildings & 0.092 & 0.092 & \textbf{0.169} & 0.095 & 0.126 & 0.172 & 0.909 & 0.931 & 0.792 & 0.862 & 0.965 & 0.955 & 0.896 & 0.932 \\
iSAID vehicles$^{*}$ & 0.164 & 0.293 & \textbf{0.304} & 0.293 & \textbf{0.304} & 0.304 & 0.939 & 0.929 & 0.918 & 0.918 & 0.972 & 0.955 & 0.950 & 0.950 \\
\emph{mean, aerial} & 0.128 & 0.192 & \textbf{0.237} & 0.194 & 0.215 & 0.238 & 0.924 & 0.930 & 0.855 & 0.890 & 0.969 & 0.955 & 0.923 & 0.941 \\
\midrule
BBBC038 nuclei & \textbf{0.842} & 0.000 & 0.726 & \textbf{0.842} & \textbf{0.842} & 0.742 & 0.197 & 1.000 & 0.409 & 0.197 & 0.343 & 1.000 & 0.522 & 0.343 \\
LIVECell 8 lines & \textbf{0.216} & 0.000 & 0.000 & \textbf{0.216} & \textbf{0.216} & 0.078 & 0.696 & 1.000 & 1.000 & 0.696 & 0.806 & 1.000 & 1.000 & 0.806 \\
LIVECell A172 & 0.481 & 0.503 & \textbf{0.578} & 0.503 & \textbf{0.578} & 0.586 & 0.504 & 0.647 & 0.545 & 0.545 & 0.771 & 0.730 & 0.673 & 0.673 \\
LIVECell SH-SY5Y & 0.236 & 0.257 & \textbf{0.375} & 0.252 & \textbf{0.375} & 0.375 & 0.807 & 0.849 & 0.729 & 0.729 & 0.937 & 0.901 & 0.845 & 0.845 \\
LIVECell micro-SAM split & 0.390 & 0.336 & \textbf{0.420} & 0.383 & 0.410 & 0.424 & 0.686 & 0.794 & 0.706 & 0.691 & 0.822 & 0.834 & 0.781 & 0.804 \\
\emph{mean, microscopy} & 0.433 & 0.219 & 0.420 & 0.439 & \textbf{0.484} & 0.441 & 0.578 & 0.858 & 0.678 & 0.572 & 0.736 & 0.893 & 0.764 & 0.694 \\
\midrule
\emph{mean, all 13} & 0.239 & 0.404 & 0.495 & 0.475 & \textbf{0.503} & 0.511 & 0.670 & 0.617 & 0.545 & 0.528 & 0.784 & 0.733 & 0.674 & 0.662 \\
\bottomrule
\end{tabular}
\end{table*}

\subsection{Paired bootstrap}
\label{sec:sup_bootstrap}
A test over five projects cannot go below $p = 0.03125$, so we also resample evaluation images.
For each project we draw $10{,}000$ resamples of its $120$ evaluation images, apply the same draw to every method, and recompute pooled instance macro-F1 and the drawn fraction $D$ from per-image counts; our score is the mean over the three seeds inside each resample.
The five-project mean resamples each project independently.
$p$ is one-sided, the share of resamples in which ours is not better.

\Cref{tab:bootstrap} gives the result.
On the five microscopy projects, \snaphead leads SAM~3 with its class name by $0.214$ F1 (95\% CI $[0.195, 0.231]$) and draws $0.280$ fewer objects by hand ($[0.262, 0.298]$), with a lower $D$ on all five projects.
Against SAM~3's oracle, the F1 difference is $-0.008$ ($[-0.024, 0.007]$), while our drawn fraction is lower by $0.031$ ($[0.012, 0.048]$, $p = 0.0004$).
On RF100-VL, \snaphead beats both SAM~3 arms on everdaynew and on stomata ($p \le 0.0004$); on floating waste the interval contains zero.
The bootstrap covers every method with stored per-image predictions; aerial-pool and floods (where the switch picks the learned proposer) and few-shot Mask~R-CNN keep the Wilcoxon test.
The router (\cref{sec:router}) is tested the same way, with each bin resampling its projects independently.
On microscopy it leads SAM~3's oracle by $0.043$ F1 ($[0.031, 0.056]$, $p < 10^{-4}$) and leads both of its inputs, \snaphead by $0.051$ and SAM~3 at its labeled operating point by $0.064$.
Over all 13 projects it comes within $0.008$ F1 of the oracle and still leads both inputs.

\begin{table*}[t]
\centering
\footnotesize
\setlength{\tabcolsep}{4pt}
\caption{Paired bootstrap over evaluation images, $N{=}20$, $10{,}000$ resamples. $\Delta$: ours minus SAM~3 for F1; SAM~3 minus ours for $D$, so a positive $\Delta$ always favors ours. CI: 95\% percentile interval. $p$: one-sided. \emph{name}: SAM~3 with its class name at threshold $0.5$ (RF100-VL: as shipped). $\dagger$: oracle, phrasing and threshold chosen on evaluation labels; on RF100-VL it is chosen on box IoU, the reported metric.}
\label{tab:bootstrap}
\begin{tabular}{@{}lc ccc ccc@{}}
\toprule
 & & \multicolumn{3}{c}{vs.\ SAM~3, name} & \multicolumn{3}{c}{vs.\ SAM~3$^\dagger$} \\
\cmidrule(lr){3-5}\cmidrule(lr){6-8}
Project & ours & $\Delta$ & CI & $p$ & $\Delta$ & CI & $p$ \\
\midrule
\multicolumn{8}{@{}l}{\emph{Microscopy, instance F1 at mask IoU $0.5$}} \\
BBBC038 nuclei & 0.842 & $+0.842$ & $[0.797, 0.882]$ & $<\!10^{-4}$ & $+0.100$ & $[0.061, 0.147]$ & $<\!10^{-4}$ \\
LIVECell 8 lines & 0.216 & $+0.216$ & $[0.192, 0.237]$ & $<\!10^{-4}$ & $+0.138$ & $[0.109, 0.162]$ & $<\!10^{-4}$ \\
LIVECell A172 & 0.481 & $-0.022$ & $[-0.065, 0.019]$ & 0.85 & $-0.104$ & $[-0.128, -0.082]$ & 1.00 \\
LIVECell SH-SY5Y & 0.236 & $-0.021$ & $[-0.059, 0.013]$ & 0.88 & $-0.139$ & $[-0.162, -0.117]$ & 1.00 \\
LIVECell micro-SAM split & 0.390 & $+0.054$ & $[-0.004, 0.103]$ & 0.03 & $-0.034$ & $[-0.085, 0.010]$ & 0.93 \\
\emph{mean of 5} & 0.433 & $+0.214$ & $[0.195, 0.231]$ & $<\!10^{-4}$ & $-0.008$ & $[-0.024, 0.007]$ & 0.86 \\
\midrule
\multicolumn{8}{@{}l}{\emph{Microscopy, drawn fraction $D$}} \\
BBBC038 nuclei & 0.197 & $+0.803$ & $[0.741, 0.858]$ & $<\!10^{-4}$ & $+0.172$ & $[0.108, 0.240]$ & $<\!10^{-4}$ \\
LIVECell 8 lines & 0.696 & $+0.304$ & $[0.261, 0.355]$ & $<\!10^{-4}$ & $+0.016$ & $[-0.015, 0.043]$ & 0.14 \\
LIVECell A172 & 0.504 & $+0.144$ & $[0.109, 0.173]$ & $<\!10^{-4}$ & $+0.027$ & $[0.004, 0.047]$ & 0.01 \\
LIVECell SH-SY5Y & 0.807 & $+0.042$ & $[0.019, 0.061]$ & 0.0001 & $-0.078$ & $[-0.103, -0.057]$ & 1.00 \\
LIVECell micro-SAM split & 0.686 & $+0.108$ & $[0.068, 0.148]$ & $<\!10^{-4}$ & $+0.015$ & $[-0.030, 0.054]$ & 0.25 \\
\emph{mean of 5} & 0.578 & $+0.280$ & $[0.262, 0.298]$ & $<\!10^{-4}$ & $+0.031$ & $[0.012, 0.048]$ & 0.0004 \\
\midrule
\multicolumn{8}{@{}l}{\emph{RF100-VL, instance F1 at box IoU $0.5$}} \\
everdaynew & 0.294 & $+0.187$ & $[0.136, 0.237]$ & $<\!10^{-4}$ & $+0.149$ & $[0.104, 0.192]$ & $<\!10^{-4}$ \\
stomata & 0.260 & $+0.260$ & $[0.220, 0.291]$ & $<\!10^{-4}$ & $+0.106$ & $[0.048, 0.151]$ & 0.0004 \\
floating waste & 0.196 & $-0.041$ & $[-0.119, 0.041]$ & 0.84 & $-0.058$ & $[-0.143, 0.030]$ & 0.90 \\
\midrule
\multicolumn{8}{@{}l}{\emph{Router (\cref{sec:router}), bin means of instance F1}} \\
microscopy (5) & 0.484 & $+0.265$ & $[0.250, 0.278]$ & $<\!10^{-4}$ & $+0.043$ & $[0.031, 0.056]$ & $<\!10^{-4}$ \\
natural (6) & 0.615 & $-0.014$ & $[-0.026, -0.004]$ & 1.00 & $-0.046$ & $[-0.054, -0.034]$ & 1.00 \\
all 13 & 0.503 & $+0.099$ & $[0.090, 0.106]$ & $<\!10^{-4}$ & $-0.008$ & $[-0.014, -0.000]$ & 0.98 \\
\midrule
 & & \multicolumn{3}{c}{vs.\ \snaphead} & \multicolumn{3}{c}{vs.\ SAM~3, labeled} \\
\cmidrule(lr){3-5}\cmidrule(lr){6-8}
\multicolumn{8}{@{}l}{\emph{Router against its two inputs, bin means of instance F1}} \\
microscopy (5) & 0.484 & $+0.051$ & $[0.044, 0.059]$ & $<\!10^{-4}$ & $+0.064$ & $[0.051, 0.077]$ & $<\!10^{-4}$ \\
natural (6) & 0.615 & $+0.501$ & $[0.477, 0.522]$ & $<\!10^{-4}$ & $-0.029$ & $[-0.034, -0.023]$ & 1.00 \\
all 13 & 0.503 & $+0.264$ & $[0.253, 0.275]$ & $<\!10^{-4}$ & $+0.008$ & $[0.002, 0.014]$ & 0.0027 \\
\bottomrule
\end{tabular}
\end{table*}

\subsection{Microscopy specialists}
\label{sec:sup_cellpose}
We run Cellpose \texttt{cyto3}~\cite{stringer2025cellpose3} zero-shot, with its own size model and no per-project tuning, on all five microscopy projects, and StarDist (\texttt{2D\_versatile\_fluo})~\cite{schmidt2018stardist} on BBBC038, the one fluorescence project.
Neither model outputs a class; on the eight-line project every Cellpose mask takes the line most common in the labeled images.
We also feed Cellpose's boxes to \snaphead as one more proposal source (\cref{sec:absorption}): each box prompts the frozen decoder, and \snaphead classifies the result, at $N{=}20$ over three seeds.

\Cref{tab:cellpose} gives the result.
Cellpose alone cannot separate the eight cell lines ($0.053$ F1); \snaphead on its proposals reaches $0.338$.
Cellpose proposals raise \snaphead on all five projects, from a mean F1 of $0.433$ to $0.544$, above SAM~3's oracle in \cref{tab:master}.
Its masks enter \snaphead as boxes re-decoded by SAM2, and the re-decoded masks match fewer cells than Cellpose's own; of those that match a cell, \snaphead keeps $80$--$95\%$ and rejects $1$--$14\%$ as background, most on SH-SY5Y and the micro-SAM split.
This is why the combination still draws more by hand than Cellpose alone ($D = 0.495$ against $0.310$).
A variant that exempts Cellpose proposals from the background reject reaches $0.560$ F1 and $D = 0.464$; the default treats every source alike.

\emph{Training overlap.}
Our LIVECell images come from LIVECell's own training annotation file, and Cellpose \texttt{cyto3} was trained on a mix that includes LIVECell; BBBC038 is in its nuclei training mix as well.
We therefore report the Cellpose and StarDist numbers here, as an upper bound for unseen cells.

\begin{table}[t]
\centering
\footnotesize
\setlength{\tabcolsep}{2pt}
\caption{Microscopy specialists, instance F1 at mask IoU $0.5$, with $D$ and $E$ as in \cref{tab:master}. \emph{Cellpose}: \texttt{cyto3}, zero-shot. \emph{SD}: StarDist, BBBC038 only. \emph{CP + head}: Cellpose boxes as proposals for \snaphead, $N{=}20$, mean over three seeds. \emph{head}: \snaphead with its own proposals (\cref{tab:master}). $\ddagger$: likely trained on these images.}
\label{tab:cellpose}
\begin{tabular}{@{}l ccc c ccc c@{}}
\toprule
 & \multicolumn{3}{c}{Cellpose$^\ddagger$} & SD$^\ddagger$ & \multicolumn{3}{c}{CP + head} & head \\
\cmidrule(lr){2-4}\cmidrule(lr){5-5}\cmidrule(lr){6-8}\cmidrule(lr){9-9}
Project & F1 & $D$ & $E$ & F1 & F1 & $D$ & $E$ & F1 \\
\midrule
BBBC038 nuclei & 0.817 & 0.220 & 0.366 & 0.756 & 0.845 & 0.180 & 0.334 & 0.842 \\
8 lines & 0.053 & 0.257 & 0.365 & --- & 0.338 & 0.542 & 0.668 & 0.216 \\
A172 & 0.751 & 0.335 & 0.514 & --- & 0.626 & 0.453 & 0.680 & 0.481 \\
SH-SY5Y & 0.668 & 0.444 & 0.631 & --- & 0.367 & 0.719 & 0.868 & 0.236 \\
micro-SAM split & 0.799 & 0.293 & 0.446 & --- & 0.545 & 0.581 & 0.727 & 0.390 \\
\midrule
\emph{mean of 5} & 0.618 & 0.310 & 0.464 & --- & 0.544 & 0.495 & 0.655 & 0.433 \\
\bottomrule
\end{tabular}
\end{table}

\subsection{Descriptor halves}
\label{sec:sup_halves}
The descriptor $x$ joins two $256$-d halves (\cref{eq:descriptor}).
We refit \snaphead on each half alone under the given-masks protocol of \cref{tab:naming} (\cref{tab:halves}).
The joint descriptor has the highest $\kappa$ at every budget on the eight lines, and on the A172 size split from two images on; at one image all three are near chance there.
The halves carry different information.
On the eight lines the mask-pooled half carries most of the signal ($0.747$ against $0.754$ at $60$ images) and the pointer half alone reaches $0.501$.
On the size split the pointer half leads the mask-pooled half at two and five images, and at $60$ images each half alone trails $x$ ($0.641$ and $0.595$ against $0.722$).

\begin{table}[t]
\centering
\footnotesize
\setlength{\tabcolsep}{4pt}
\caption{Descriptor halves, given ground-truth masks as in \cref{tab:naming}. Cohen's $\kappa$, mean over three seeds. \emph{images}: labeled images; on the eight lines the budget is set per line, and $20$ per line takes the whole $60$-image pool. Bold: best per row.}
\label{tab:halves}
\begin{tabular}{@{}lc ccc@{}}
\toprule
Project & images & $x$ & mask-pooled & pointer \\
\midrule
8 lines & 8 & \textbf{0.429} & 0.417 & 0.344 \\
 & 16 & \textbf{0.558} & 0.553 & 0.421 \\
 & 40 & \textbf{0.713} & 0.708 & 0.479 \\
 & 60 & \textbf{0.754} & 0.747 & 0.501 \\
\midrule
A172, size split & 1 & 0.068 & \textbf{0.110} & 0.084 \\
 & 2 & \textbf{0.206} & 0.163 & 0.193 \\
 & 5 & \textbf{0.554} & 0.455 & 0.515 \\
 & 20 & \textbf{0.674} & 0.597 & 0.571 \\
 & 60 & \textbf{0.722} & 0.641 & 0.595 \\
\bottomrule
\end{tabular}
\end{table}

\subsection{Per-dataset alternatives}
\label{sec:sup_ablations}

\Cref{tab:sup_ablations} lists alternatives to our fixed configuration and what each changes. Recall is per-image recall at IoU $0.5$ on the evaluation images; F1, mean instance F1 over the 13 projects at $N{=}20$.

\begin{table*}[t]
\centering
\footnotesize
\caption{Alternatives to the fixed configuration and their trade-offs.}
\label{tab:sup_ablations}
\begin{tabular}{@{}p{0.2\linewidth}p{0.42\linewidth}p{0.32\linewidth}@{}}
\toprule
alternative & result & trade-off \\
\midrule
AMG knobs tuned from seed images & On LIVECell-A172 the costliest grid reaches recall $0.358$ at $84.9$\,s per image; the learned proposer reaches $0.575$ at $0.41$\,s. & More compute buys less recall than our proposer. \\
AMG filters switched off & The raw pool covers $0.694$ of A172 cells; SAM2's predicted-IoU $>0.7$ filter cuts that to $0.146$, and NMS removes almost nothing. & About $10^4$ masks per image is too many to score. \\
AMG filter learned from seed images & On single-scale grids it keeps $78$--$99\%$ of the raw pool on cells (A172 $0.355$ of $0.369$), but fails on crops and COCO vehicles. & Still below the learned proposer ($0.575$ on A172) at higher cost. \\
Training-free top-hat blob proposer & Joined with AMG it lifts A172 recall $0.167{\to}0.282$ and 8type $0.259{\to}0.330$; alone it finds almost nothing large ($0.014$ on VOC). & The learned proposer is better wherever top-hat helps. \\
Native-resolution tiling, aerial tiles & $83\%$ of iSAID vehicles are under $8$\,px after resize, where both methods find about $1\%$; above it we match SAM~3. Tiled AMG costs about $110$\,s. & It lifts SAM~3 and us alike, so we treat it as preprocessing. \\
Proposer choice by cross-validation & An oracle picking the best proposer set per project beats the one-line switch by only $0.005$ ($N{=}5$) and $0.002$ ($N{=}20$) mean F1. & The one-line switch already captures the gain. \\
Fragments trained as background & Mean F1 falls $0.239{\to}0.233$, and the share left to draw rises. & A fragment's descriptor looks like the whole object, so \snaphead rejects whole objects. \\
Learned ranker over same-location versions & Mean F1 is indistinguishable from the plain rule at three decimals ($0.239$ either way), but loses more than seed noise on livecell-A172. & The plain rule reaches the same mean without that loss; its benefit is in effort (\cref{sec:sup_guarantee}). \\
Distance-map proposer, a plug-in source for dense cells & At $N{=}5$ on LIVECell-A172 it reaches $0.463 \pm 0.013$ F1, above the center head's $0.433 \pm 0.046$, with a third of the spread. & Its gain is project-specific: on BBBC038 two of three seeds give no usable head ($0.222 \pm 0.313$), and it trails in recall on every other project measured. As a pool source (\cref{sec:absorption}) it can be switched on from the labels, as \cref{eq:gate} switches the learned proposer. \\
\bottomrule
\end{tabular}
\end{table*}

We use containment-aware suppression instead: keeping one object per location, using containment as well as IoU. \Cref{sec:sup_containment} reports its effect on F1 and duplicate detections.

\section{Annotation Effort}
\label{sec:sup_effort}

\subsection{Effort-optimal dial}
\label{sec:sup_guarantee}

Minimizing $c_{\mathrm{delete}}\mathrm{FP} + c_{\mathrm{draw}}\mathrm{FN}$ weights errors by
$c_{\mathrm{draw}}/c_{\mathrm{delete}}$; maximizing $F_\beta$ weights them by $\beta^2$. They
agree when
\begin{equation}
\beta = \sqrt{c_{\mathrm{draw}}/c_{\mathrm{delete}}}.
\label{eq:beta}
\end{equation}
Our default $F_1$ dial assumes deleting a false positive costs as much as drawing a
missed object.

Eq.~\ref{eq:effort} bounds the dial at manual cost on its tuning evidence. We retuned it for
\cref{sec:effort}'s check/refine/draw vector: the bound holds on $231/231$ tuning curves,
$231/231$ cross-validated replays, and $221/231$ held-out splits ($95.7\%$). The ten
violations are small ($0.02\%$--$8.0\%$ over manual) and confined to $N{\le}10$ on
three projects (coco-kitchen, coco-vehicles, lvis-rare20).

Over $0 < w_{\mathrm{check}} \le w_{\mathrm{refine}} < 1$, \paper beats manual at the
default weights on all 13 projects. The winning region ranges from $0.26$ on lvis-rare20 to
$0.99$ on bbbc038-nuclei.

\paragraph{Containment.}
\label{sec:sup_containment}
\cref{sec:sup_ablations} studies a one-object-per-location suppression rule. With our descriptor, F1 is
barely changed ($0.239$ without it, $0.237$ with it, over 13 projects at $N{=}20$). We remove
about $600$ duplicate detections per project. The effort gain grows with the check weight:
break-even sits at $w_{\mathrm{check}} \approx 0.259$ ($w_{\mathrm{refine}}=0$); past that, we
win over $64\%$ of the cited region.

\paragraph{Sensitivity.}
\label{sec:sup_sensitivity}
We recounted stored action counts across all 15 methods under three alternative policies: a
looser accept/refine line, a loose match charged as a full redraw, and a required mask fix
with the class fix. Order stability is $0.857$, top-1 stability $0.889$ (Kendall $\tau$ mean
$0.892$, 7560 cells). The harshest policy, charging both corrections as a full
redraw, moves the ranking most.

\subsection{Effort across the cost plane}
\label{sec:sup_effort_baselines}

\begin{figure}[t]
\centering
\includegraphics[width=\linewidth]{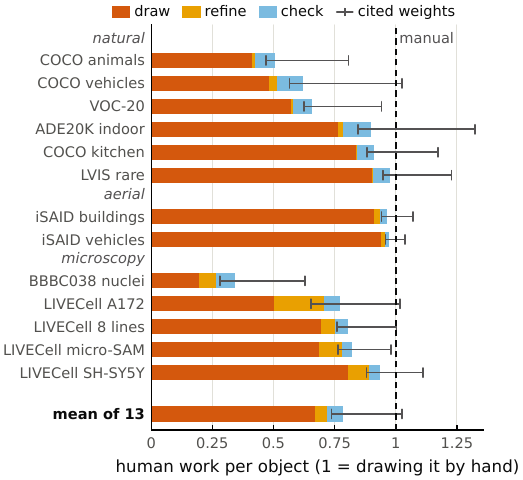}
\caption{Effort per project at $N{=}20$, split into drawn ($D$), refined ($w_{\mathrm{refine}}R$) and checked ($w_{\mathrm{check}}C$) shares. Whiskers: cited ranges, $w_{\mathrm{check}}$ $0.063$--$0.473$~\cite{papadopoulos2016verify,kuznetsova2020open} and $w_{\mathrm{refine}}$ $0.268$--$0.423$~\cite{oehme2026aramsam,tyagi2023boxcell}, extended to our $0.5$ default. Dashed: manual.}
\label{fig:effort}
\end{figure}

We split the effort of \cref{sec:loop} by project at $N{=}20$ in \cref{fig:effort}.

The cited cost ratios bound $w_{\mathrm{check}}$ at
$0.063$--$0.473$~\cite{papadopoulos2016verify,su2012crowdsourcing,kuznetsova2020open} and
$w_{\mathrm{refine}}$ at $0.268$--$0.423$~\cite{oehme2026aramsam,tyagi2023boxcell}.

\subsection{Annotation loop}
\label{sec:sup_loop}

We replay \paper's loop: $36$ rounds of five images over a $180$-image pool. Each round,
the annotator corrects \paper's predictions and we refit \snaphead. The configuration matches
\cref{tab:config}; the proposer switch fires on LIVECell-A172 from the first refit and on none
of the other three projects.

\cref{tab:loop} gives the endpoint. On A172, AMG alone would leave $0.915$ to draw; the learned proposer
drives the drop. Cumulative effort falls below manual from the second round on, staying below
it over $0.988$--$1.000$ of the cited range.

\begin{table}[!t]
\centering
\footnotesize
\setlength{\tabcolsep}{4pt}
\caption{Endpoint of the $36$-round loop, \paper configuration, mean over three seeds. $D$: hand-drawn share, $\pm$ sd. $E$: effort at default weights. \emph{range}: share of the cited range below manual.}
\label{tab:loop}
\begin{tabular}{@{}lrrrr@{}}
\toprule
Project & $n_{\mathrm{GT}}$ & $D$\,$\downarrow$ & $E$\,$\downarrow$ & range\,$\uparrow$ \\
\midrule
bbbc038-nuclei & 6951 & $0.213 \pm 0.006$ & 0.357 & 1.000 \\
coco-animals & 452 & $0.324 \pm 0.014$ & 0.453 & 1.000 \\
coco-vehicles & 687 & $0.455 \pm 0.002$ & 0.602 & 0.988 \\
livecell-A172 & 38500 & $0.504 \pm 0.003$ & 0.767 & 0.994 \\
\bottomrule
\end{tabular}
\end{table}

\section{Runtime and Memory Cost}
\label{sec:sup_cost}

We measured every number here on \paper's code paths: one H100 for GPU stages, an Intel Xeon
Platinum 8480+ for CPU work, and headless Chrome 153 (onnxruntime-web, WASM, one thread) for
the browser.

\begin{table}[!t]
\centering
\footnotesize
\setlength{\tabcolsep}{2.5pt}
\caption{Measured per-image cost, median over 8 images; AMG at production density. \emph{stage only}: proposer vs.\ AMG alone. \emph{deployed path}: adds decode, NMS, re-prompt and descriptor. Bold: deployed learned proposer. Accuracy: \cref{sec:proposer_res}.}
\label{tab:cost}
\begin{tabular}{lr}
\toprule
Stage & measured \\
\midrule
encoder pass, once per image & $18.6$--$19.7$\,ms \\
encoder peak VRAM & $1.64$\,GB \\
production AMG proposals & $9.0$--$9.9$\,s/img \\
AMG at half grid density & $0.53$--$0.75$\,s/img \\
learned proposer, stage only & $0.02$--$0.49$\,s/img \\
learned proposer, parameters & $97{,}603$ \\
learned proposer, training & $1.5$--$3.9$\,s \\
descriptor pooling over proposals & $0.14$--$0.88$\,s/img \\
\midrule
\snaphead fit & $8$\,ms--$2.6$\,s \\
grouped CV, backgrounded (feeds the dial) & $0.08$--$304$\,s \\
\snaphead artifact, 1 class / 10 classes (512-d) & $33$ / $143$\,KB \\
\midrule
proposer fit in \paper (3 images, 300 steps) & $2.98$\,s \\
\textbf{learned rows, deployed path} & $\mathbf{1.68}$\,\textbf{s/img} \\
peak CUDA, union mode & $3.06$\,GB \\
supplied zero-shot boxes: decode $+$ descriptor & $0.81$\,s/img \\
\midrule
browser decode per prompt, median & $147$--$156$\,ms \\
browser decode per prompt, p95 & $171$--$182$\,ms \\
decoder ONNX, session create & $15.8$\,MiB, $466$--$507$\,ms \\
embedding frame, parse & $8.00$\,MiB, $42$--$56$\,ms \\
\bottomrule
\end{tabular}
\end{table}

\begin{figure}[t]
\centering
\includegraphics[width=\linewidth]{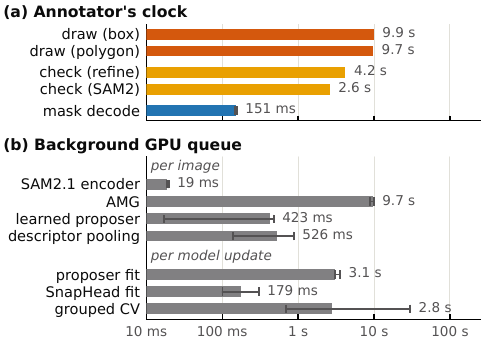}
\caption{Time on the annotator's clock and in the background. (a)~Seconds per object to draw or check~\cite{tyagi2023boxcell,oehme2026aramsam}, plus the browser's decode wait (one CPU thread). (b)~GPU stages per image or update, one H100. Bars: median; whiskers: range; log axis.}
\label{fig:time}
\end{figure}

\paragraph{Per-image cost.}
In \cref{tab:cost}, the encoder stays flat across a $20{\times}$ range of input
pixels, since every image resizes to one square before the trunk. Production AMG, by
contrast, costs roughly $480{\times}$ more and scales with proposal count, not input size.
Halving the AMG grid cuts that cost $13$--$17{\times}$ while the encoder stays unchanged to
$0.2$\,ms. Cost scales with the square of the AMG point-grid density, so we state that
density for every AMG figure.

\paragraph{\snaphead size.}
Its artifact is $33$--$143$\,KB at $D{=}512$ (one to ten classes). It is bounded
by the ceiling $(C{+}1)(D{+}1)+2D$ for $C$ classes; the weights make up $95$--$98\%$ of the
serialized bytes.

\begin{figure}[t]
\centering
\includegraphics[width=\linewidth]{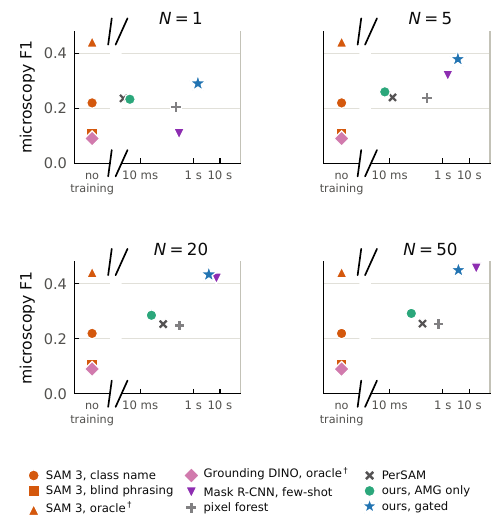}
\caption{Mean instance F1 over the five microscopy projects against time to adapt, per budget $N$ (log $x$-axis). \emph{no training}: prompt-only methods. \emph{ours, gated}: \snaphead fit (CPU) plus proposer training (GPU, only where the gate fires), as in \cref{tab:master}.}
\label{fig:adapt}
\end{figure}

\paragraph{Adaptation time.}
\label{sec:sup_adapt}
Among methods that learn only the class, \snaphead fits in $0.03$\,s (CPU) against $0.07$\,s
for PerSAM and $0.30$\,s for the pixel forest, at a higher F1 ($0.285$ vs.\ $0.253$/$0.249$).
Among methods that also learn object location, \emph{ours, gated} adapts in $3.83$\,s against
$7.39$\,s for Mask~R-CNN, again at a higher F1 ($0.433$ vs.\ $0.419$).

Our proposer trains a fixed $300$ steps, so cost stays flat as labels grow: $3.70$--$3.81$\,s
from $N{=}5$ to $50$ on A172, against $1.41$--$15.88$\,s for Mask~R-CNN. \cref{fig:adapt}
pairs this cost with \cref{sec:master}'s accuracy: \snaphead fits under $0.1$\,s
(median $0.004$\,s at $N{=}1$, $0.066$\,s at $N{=}50$), and \emph{ours, gated} stays under
$4$\,s.

An update lands right after \snaphead's fit: on the five microscopy projects at $N{=}20$,
$0.04$--$0.15$\,s (median $0.04$\,s). The optional dial's
cross-validation runs afterward in the background, without blocking predictions:
$0.8$--$54.7$\,s later (median $23.6$\,s) on the same five cells.

\paragraph{Browser decoder.}
A prompt decodes locally in $147$--$156$\,ms (median) and $171$--$182$\,ms (p95),
single-threaded on the CPU above. Box and point prompts cost the same. Over 50 box prompts per image, the displayed mask agrees with the
server's torch decoder at mean IoU $0.993$--$0.997$, never below $0.972$.

\end{document}